\documentclass[10pt,twocolumn,letterpaper]{article}

\usepackage[pagenumbers]{cvpr} 

\usepackage{ulem}
\usepackage{amsmath}
\usepackage{algorithm}
\usepackage{algpseudocode}   

\usepackage{graphicx}
\usepackage{caption}
\usepackage{adjustbox}
\usepackage{arydshln}

\renewcommand{\topfraction}{0.9}
\renewcommand{\bottomfraction}{0.8}
\renewcommand{\textfraction}{0.05}
\renewcommand{\floatpagefraction}{0.99}

\definecolor{cvprblue}{rgb}{0.21,0.49,0.74}
\usepackage{hyperref}
\hypersetup{breaklinks,colorlinks,allcolors=cvprblue}
\title{Diffusion-Based Attention Warping for Consistent 3D Scene Editing}

\author{
Eyal Gomel\qquad Lior Wolf\\
Tel-Aviv University
}

\begin{document}
\maketitle
\begin{abstract}
We present a novel method for 3D scene editing using diffusion models, designed to ensure view consistency and realism across perspectives. Our approach leverages attention features extracted from a single reference image to define the intended edits. These features are warped across multiple views by aligning them with scene geometry derived from Gaussian splatting depth estimates. Injecting these warped features into other viewpoints enables coherent propagation of edits, achieving high fidelity and spatial alignment in 3D space. Extensive evaluations demonstrate the effectiveness of our method in generating versatile edits of 3D scenes, significantly advancing the capabilities of scene manipulation compared to the existing methods. Project page: \small\url{https://attention-warp.github.io}
\end{abstract}

\section{Introduction}
\label{sec:intro}

Recent advances in diffusion models have revolutionized the landscape of 2D image editing, demonstrating unprecedented capabilities in image editing, style transfer and image inpainting~\cite{rombach2022high,Dhariwal2021DiffusionMB,song2020denoising,ho2020ddpm,zhang2023inversion,deng2022stytr2,lugmayr2022repaint}. While these achievements have solidified the position of diffusion models as the de facto standard for image editing tasks, extending such capabilities to 3D scene editing presents unique challenges. Several recent contributions have attempted to bridge this gap by applying 2D diffusion models to multiple views of a 3D scene, typically utilizing the same views employed in the scene's reconstruction~\cite{instructnerf2023,igs2gs,gaussctrl2024}. While editing based on a single view proves inadequate for comprehensive 3D manipulation, the simultaneous editing of multiple views introduces significant challenges in maintaining edit consistency across perspectives.

Prior approaches have addressed the consistency challenge through various mechanisms, predominantly focusing on information propagation between views~\cite{dong2024vica, chen2024dge, wang2025view, khalid2023latenteditor}. However, such approaches frequently result in a loss of edit fidelity and conceptual clarity, as the attempt to reconcile potentially conflicting information from multiple views leads to blurred or compromised results. 

In contrast, we propose a novel paradigm that leverages only a single-view edit as the primary manipulation source, then systematically projects the edit attention feature maps onto other views using the underlying 3D scene structure.
This innovative use of attention warping ensures that edits are propagated consistently across different perspectives without processing multiple frames simultaneously, significantly reducing computational complexity. 

A key innovation in our method is the incorporation of a geometry-guided warping mechanism that utilizes the depth and structural information of the scene to accurately map edits across views, maintaining spatial coherence and alignment with the 3D scene’s structure. Additionally, we propose masking and blending techniques that exploit Gaussian splatting properties, such as Gaussian normal vectors, to prevent warping to occluded or misaligned regions. These techniques ensure smooth transitions and consistency across views, refining the edit quality and preserving realistic integration throughout the 3D model. Our contributions enable high-quality, multi-view-consistent 3D edits that are computationally efficient and robust.

The effectiveness of our approach is demonstrated through extensive experimental validation across a diverse range of editing scenarios and scene types. Our method consistently outperforms existing approaches in terms of edit quality, spatial consistency, and semantic fidelity, as verified through both quantitative metrics and user studies.

\section{Related Work}
\label{sec:related}
Our method builds upon advancements in image-text-based editing using diffusion models ~\cite{song2020denoising, rombach2022high, brooks2022instructpix2pix, zhang2023adding}, Gaussian Splatting~\cite{kerbl3Dgaussians, Huang2DGS2024} representations, and 3D scene editing techniques~\cite{instructnerf2023,igs2gs, chen2024gaussianeditor, wang2024gaussianeditor, chen2024dge, dong2024vica, wang2025view, khalid2023latenteditor}. We review these key areas and highlight how our approach differs from existing methods.

\noindent{\bf Text Based Image Editing w/ Diffusion Models\quad}
Diffusion models have become essential for text-guided image editing by leveraging a noise-adding and denoising process to modify images with precision. This process involves iteratively refining a noisy image $\mathbf{x}_T$ across timesteps $t$ using learned noise predictions $\epsilon_\theta(\mathbf{x}_t, \mathbf{t})$.

Central to diffusion models, especially in text-guided applications, are self-attention and cross-attention mechanisms within its U-Net structure~\cite{ronneberger2015u}. Self-attention captures dependencies across different regions within the image, enhancing coherence and structure during generation. Cross-attention incorporates external guidance, such as text prompts or conditioning images, by mapping relationships between image features and the conditioning input.

{InstructPix2Pix}~\cite{brooks2022instructpix2pix} is an image-to-image translation method built upon {Stable Diffusion}~\cite{rombach2022high}. It fine-tunes Stable Diffusion using synthetic instruction data, enabling the model to perform targeted image editing based on both textual prompts and reference images for structural alignment. This approach is part of a broader class of image-to-image translation methods~\cite{meng2021sdedit, bar2022text2live, hertz2022prompt, tumanyan2023plug, mokady2023null, parmar2023zero, kawar2023imagic}.

{ControlNet}~\cite{zhang2023adding} enhances the general diffusion framework by introducing trainable auxiliary control structures that condition the generation process on additional inputs, such as depth maps, edges, and segmentation masks. Unlike traditional cross-attention conditioning, ControlNet incorporates a parallel trainable path that merges with the original diffusion model.

Our method extends these approaches by editing the content guided by attention mechanisms and also warping the self and cross-attention feature maps across views. This attention warping propagates edits consistently in 3D scenes, ensuring alignment across different viewpoints. This unique handling of attention feature maps sets our method apart by addressing multi-view consistency challenges more effectively than traditional 2D-based approaches.

\noindent{\bf 3D Scene Representation\quad}
Neural Radiance Fields (NeRF)~\cite{mildenhall2021nerf} and derivative works~\cite{wang2021neus,liu2020neural,muller2022instant,barron2021mip,barron2022mip} have opened up new possibilities in computer vision and computer graphics, including 3D scene reconstruction, editing, segmentation, etc. These models represent scenes as continuous 3D functions, synthesizing novel views by learning volumetric density and color at each 3D point.

3D Gaussian Splatting~\cite{kerbl3Dgaussians} (3DGS) represents scenes using discrete 3D Gaussian elements, enabling efficient rendering while capturing complex details. Some methods build upon this representation to enhance depth and normal consistency, such as {2DGS}~\cite{Huang2DGS2024} which improves 3DGS by collapsing the 3D volume into 2D oriented planar Gaussian disks, ensuring view-consistent geometry and intrinsic surface modeling. 2DGS employs a perspective-accurate splatting process using ray-splat intersections. Depth distortion and normal consistency regularization further enhance the reconstruction quality, supporting detailed geometry. 

Our approach leverages the 2DGS representation to facilitate scene edits, incorporating attention-based diffusion models to modify the splats' attributes while preserving scene integrity.

\noindent{\bf 3D Scene Editing with Diffusion Models\quad}
3D scene editing and stylization are pivotal in computer vision, enabling diverse applications in neural radiance fields. Approaches like ~\cite{zheng2023editablenerf,lazova2023control,kania2022conerf} offer controllable scene modifications that adjust geometry and appearance. Methods such as ~\cite{huang2022stylizednerf,zhang2022arf,nguyen2022snerf} achieve 3D-consistent stylizations by incorporating mutual 2D-3D learning, while techniques like ~\cite{wu2022palettenerf,gong2023recolornerf} focus specifically on color manipulation.

Leveraging image-text models~\cite{radford2021learningtransferablevisualmodels,dosovitskiy2020image} for guiding 3D generation has been explored in works such as~\cite{jain2022zero, wang2022clip, wang2023nerf}. DreamFusion~\cite{poole2022dreamfusion} introduced \textit{Score Distillation Sampling} (SDS), a method that utilizes gradients from diffusion models to guide 3D model updates. This approach has been adopted in several followup studies~\cite{poole2022dreamfusion, li2024focaldreamer, sella2023vox, zhuang2023dreameditor, park2023ed, cheng2023progressive3d} to apply diffusion priors for refining 3D representations and enabling complex edits.

Additionally, the technique of \textit{Iterative Dataset Update} has been proposed, with Instruct-NeRF2NeRF~\cite{instructnerf2023} introducing methods that edit individual views and update the dataset to refine 3D scene representations while maintaining coherence. Works leveraging this strategy, such as GaussianEditor~\cite{chen2024gaussianeditor, wang2024gaussianeditor}, IGS2GS~\cite{igs2gs}  and other related methods~\cite{yu2023edit, xu2023instructp2p, wang2024proteusnerf}, have shown promising results but still face limitations due to the inherent challenges of using 2D diffusion models for multiview edits and maintaining consistent geometry across views.

Achieving multiview consistency remains a significant challenge in 3D scene editing. Some methods utilize pretrained 2D models to ensure temporal coherence. Notable works, including ViCA-NeRF~\cite{dong2024vica}, DGE~\cite{chen2024dge}, GaussCtrl~\cite{gaussctrl2024}, VCEdit~\cite{wang2025view}, and LatentEditor~\cite{khalid2023latenteditor}, have proposed various approaches to address this challenge. ViCA-NeRF leverages NeRF depth information to establish pixel correspondences across views, enhancing multiview alignment. DGE, inspired by video generation and editing methods, employs 2D image generators for image sequences, editing multiple views simultaneously using spatio-temporal attention and enforcing epipolar constraints to maintain consistency. GaussCtrl utilizes ControlNet conditioned on depth to guide generation, aligning latents across multiple key views and ensuring coherence through the depth-conditioned model. VCEdit consolidates the cross-attention map space between views by leveraging pretrained Gaussians and aligns latents at each diffusion step through a finetuned copy of a Gaussian splatting model. LatentEditor focuses on local editing by optimizing NeRF in the diffusion latent space and applying a latent space mask for localized guidance. These methods represent different strategies for enhancing multiview consistency by leveraging latent representations, attention maps, and depth-based conditioning.

Despite these advancements, the current approaches have limitations:
1. Processing multiple views simultaneously restricts the application of specific edit styles to individual views.
2. Multi-view editing in diffusion models can be computationally intensive and memory-demanding.
3. Existing methods often struggle to use a single, non-diffusion-based edited image to apply consistent edits to a 3D model.

Our approach addresses these limitations by processing one image at a time through a warping mechanism that consistently propagates edits across multiple views. This design provides us the flexibility to select any edited image as the starting point for the warping process, enabling tailored edits that can be applied efficiently. By utilizing attention-based warping, we reduce computational load and memory usage while ensuring that edits are accurately reflected across the 3D model, thereby maintaining coherence and consistency in the final result.

\begin{figure}[t]
    \centering
    \includegraphics[width=\linewidth]{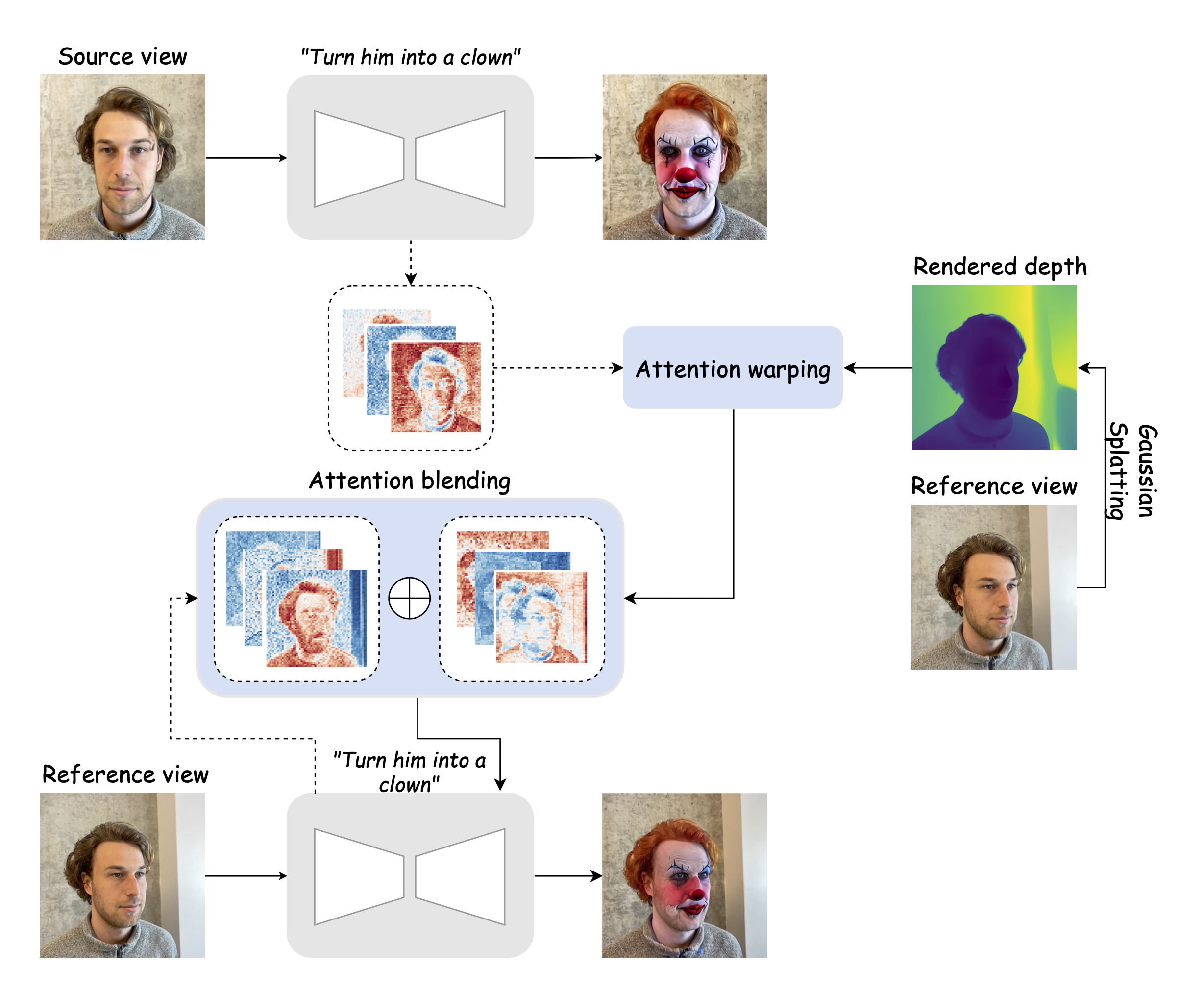}
    \caption{Overview of our method. A single source image is edited using a 2D diffusion model that is conditioned on some prompt. The attention feature maps employed during this process are saved. Given a new reference view, the maps are warped to this view based on the 3D depth map of the reference view. A diffusion model is then applied to the reference view using a blending of the attention feature maps obtained during the diffusion process itself and those that arise from the source view.\vspace{-.3cm}}
    \label{fig:method}
\end{figure}

\section{Method}

Our approach enhances 2D diffusion models with attention warping to enable consistent 3D scene editing, by ensuring that edits applied to a single view are coherently propagated across all views of the 3D scene. The method, which is illustrated in Fig.~\ref{fig:method},  comprises several key components: diffusion-based editing, attention feature map warping, occlusion handling, optional masking, and iterative optimization. Below, we detail each component and the interactions between them.

\noindent{\bf Problem Definition: Diffusion-Based 3D Scene Editing\quad}
Given a 3D scene represented by a pretrained 2D Gaussian splatting model $\mathbf{S} = \{G_i\}_{i=1}^{N}$, where each Gaussian $G_i$ is parameterized by its position, orientation, scale, opacity, and color, we perform scene editing guided by textual instructions, an unedited image, and either a reference image or a depth map. Specifically, we utilize two types of diffusion models:
\textbf{InstructPix2Pix}~\cite{brooks2022instructpix2pix}: Takes a textual instruction $\mathbf{t}$ and a guided reference image $I_\text{G}$ to generate edited images. 
\textbf{ControlNet}~\cite{zhang2023adding}: Takes a textual instruction $\mathbf{t}$ and depth map $\mathbf{D}$ to guide the editing process.

Depending on the chosen model, the input to the diffusion process is either $(\mathbf{I}, \mathbf{t}, {I_\text{G}})$ or $(\mathbf{I}, \mathbf{t}, \mathbf{D})$. The diffusion model generates edited images $\mathbf{I}'$ based on these inputs, which are then used to fine-tune the Gaussian splatting representation $\mathbf{S}$.

\noindent{\bf Source View Editing and Attention Feature Map Computation\quad}

Select a \textit{source view} $\mathbf{I}_{\mathrm{src}}$ from the training set to apply the initial edit. The diffusion model processes $\mathbf{I}_{\mathrm{src}}$ along with the corresponding instruction to produce an edited image $\mathbf{I}'_{\mathrm{src}}$. During this editing process, attention feature maps $\mathbf{F}_{\mathrm{src}} = \{\mathbf{F}_{\mathrm{src}}^l\}_{l=1}^{L}$ are computed at each layer $l$ of the diffusion model, capturing the regions of the image that are influenced by the edit. Each attention feature map $\mathbf{F}_{\mathrm{src}}^l$ comprises both self-attention and cross-attention components:

\begin{equation}
    \mathbf{F}_{\mathrm{src}}^l = \{\mathbf{F}_{\mathrm{self}}^l, \mathbf{F}_{\mathrm{cross}}^l\}\,,\quad\text{where:}
\end{equation}
%
    $\mathbf{F}_{\mathrm{self}}^l = \text{SelfAttention}\left( \mathbf{Q}, \mathbf{K}, \mathbf{V} \right)$ captures the internal relationships within the source view $\mathbf{I}_{\mathrm{src}}$. 
    $\mathbf{F}_{\mathrm{cross}}^l = \text{CrossAttention}\left( \mathbf{Q}_{\mathrm{cross}}, \mathbf{K}_{\mathrm{cross}}, \mathbf{V}_{\mathrm{cross}} \right)$ integrates contextual information from external sources or conditioning inputs, which in our case is the textual prompt $\mathbf{t}$.

\noindent\textbf{Selected Attention Feature Maps} To focus on detailed and spatially fine-grained edits, we utilize attention feature maps exclusively from the up-sampling blocks at high resolutions (32 and 64). This selection ensures that the most relevant and high-resolution attention information is propagated during the warping process.

Editing both the self- and the cross-attention allows $\mathbf{F}_{\mathrm{src}}^l$ to encapsulate both the internal dynamics of the source view and the influence of external context, facilitating more coherent and context-aware edits.

\noindent{\bf Attention Feature Map Warping to Target Views\quad} To ensure consistency across different views, the maps from the source view are warped to target views. This warping leverages depth information and camera transformations. For a target view $\mathbf{I}_{\mathrm{tgt}}$, the warped feature maps $\mathbf{F}_{\mathrm{warp}}^{\mathrm{tgt}} = \{\mathbf{F}_{\mathrm{warp}}^{l,\mathrm{tgt}}\}_{l=1}^{L}$ are obtained as follows:

\begin{equation}
    \mathbf{F}_{\mathrm{warp}}^{l,\mathrm{tgt}} = \mathcal{W}\left(\mathbf{F}_{\mathrm{src}}^l, \mathbf{D}_{\mathrm{tgt}}, \mathbf{T}_{\mathrm{src}}, \mathbf{T}_{\mathrm{tgt}}\right)\,,
\label{eq:warp}
\end{equation}
\noindent where $\mathcal{W}$ denotes the warping function that utilizes the depth map $\mathbf{D}_{\mathrm{tgt}}$, along with the camera transformation matrices $\mathbf{T}_{\mathrm{src}}$, $\mathbf{T}_{\mathrm{tgt}}$, to map the attention from the source view to the target view.

The depth map is obtained from the Gaussian splitting model $\mathbf{S}$. To ensure consistency between views, we compute the normal vectors for each Gaussian in both source and target views, denoted as $\mathbf{n}^{\mathrm{src}}_i$ and $\mathbf{n}^{\mathrm{tgt}}_i$, respectively. This helps manage the differing appearances of objects from various angles. These Gaussians are used to compute the depth map $\mathbf{D}_{\mathrm{tgt}}$, which serves as an input to Equation~\ref{eq:warp}, utilized to determine the warp visibility mask. Gaussians with normal angle differences exceeding a threshold $\theta_{\mathrm{max}}=60^\circ$, i.e, the cases in which  $\mathbf{n}^{\mathrm{src}}_i \cdot \mathbf{n}^{\mathrm{tgt}}_i \geq \cos(\theta_{\mathrm{max}})$,  are excluded from the depth rendering process.

Computing the depth based only on Gaussians with similar orientations helps to ensure that only reliable Gaussians contribute to the attention warping, reducing artifacts from mismatched geometries and enhancing the accuracy of the visibility mask. 

The camera transformations \( \mathbf{T} \) are obtained by combining various components, including the intrinsic camera parameters \( \mathbf{K} \): the focal lengths \( f_x \) and \( f_y \), and the principal point offsets  \( c_x \) and \( c_y \), as well as the 3D rotation matrix \( \mathbf{R} \). 
The warping process aligns a target image to a source view by projecting 3D coordinates from the target to the source image plane and is given here completeness. First, pixel coordinates \(\mathbf{p}_{\mathrm{tgt}}\) from the target image are unprojected into 3D space using the depth map \(\mathbf{D}_{\mathrm{tgt}}\) and intrinsic matrix \(\mathbf{K}_{\mathrm{tgt}}\):
$\mathbf{P}^c_{\mathrm{tgt}} = (\mathbf{K}_{\mathrm{tgt}}^{-1} \mathbf{p}_{\mathrm{tgt}}^\mathrm{T} \mathbf{D}_{\mathrm{tgt}})^\mathrm{T}$.

The 3D point \(\mathbf{P}^c_{\mathrm{tgt}}\) is converted to world coordinates using the extrinsic matrix \(\mathbf{R}_{\mathrm{tgt}}\):
$\mathbf{P} = \mathbf{R}_{\mathrm{tgt}}^{-1} (\mathbf{P}^c_{\mathrm{tgt}})_h$,
where \(h\) denotes homogeneous coordinates. The world coordinates \(\mathbf{P}\) are transformed to the source camera's coordinates using \(\mathbf{R}_{\mathrm{src}}\):
$\mathbf{P}^c_{\mathrm{src}} = (\mathbf{R}_{\mathrm{src}} \mathbf{P}_h^\mathrm{T})^\mathrm{T}$.

The  2D pixel coordinates in the source view are:
\[
u = \frac{f_x \mathbf{P}^c_{\mathrm{src}, x}}{\mathbf{P}^c_{\mathrm{src}, z}} + c_x, \quad v = \frac{f_y \mathbf{P}^c_{\mathrm{src}, y}}{\mathbf{P}^c_{\mathrm{src}, z}} + c_y
\]

To identify out-of-bounds regions, a mask \(\mathbf{M}\) is defined:
\begin{equation}
\mathbf{M}(u, v) = 
\begin{cases} 
1, & \text{if } 0 \leq u < W \text{ and } 0 \leq v < H \\
0, & \text{otherwise}
\end{cases}\,,
\label{eq:warp_valid_mask}
\end{equation}
where \(W\) and \(H\) are the target image width and height.

The warping process applies uniformly to both self-attention and cross-attention feature maps within $\mathbf{F}_{\mathrm{src}}^l$, ensuring that all relevant attention information is accurately propagated across views.

The warped attention feature maps $\mathbf{F}_{\mathrm{warp}}^{l,\mathrm{tgt}}$ serve as an additional input of the diffusion model to guide the target view editing:
\begin{equation}
    \mathbf{I}'_{\mathrm{tgt}} = \mathrm{DM}\left(\mathbf{I}_{\mathrm{tgt}}, \mathbf{x}_{\mathrm{tgt}}, \mathbf{t}, \mathbf{F}_{\mathrm{warp}}^{\mathrm{tgt}}\right)\,,
\end{equation}
\noindent where $\mathrm{DM}$ is the diffusion model, $\mathbf{I}_{\mathrm{tgt}}$ is the target view image, $\mathbf{x}_{\mathrm{tgt}}$ can be either the guided image $I_\text{G}^{\mathrm{tgt}}$ or the depth map $\mathbf{D}_{\mathrm{tgt}}$, $\mathbf{t}$ is the textual instruction, and $\mathbf{F}_{\mathrm{warp}}^{\mathrm{tgt}}$ represents the set of warped attention feature maps guiding the diffusion process. The usage of the last input parameter within the diffusion model is detailed in Sec.~\ref{sec:attnindm}.

\noindent{\bf Modifying the DM Attention and Handling Occlusions\quad}
\label{sec:attnindm}
During the diffusion process for the target image, we blend the warped attention with the attention computed directly from the target view's edit, denoted as $\mathbf{F}_{\mathrm{new}}^{\mathrm{tgt}}$, as follows:
\begin{align}
    \mathbf{F}_{\mathrm{masked}}^{l,\mathrm{tgt}} &= \mathbf{F}_{\mathrm{warp}}^{l,\mathrm{tgt}} \circ \mathbf{M} + \mathbf{F}_{\mathrm{new}}^{l,\mathrm{tgt}} \circ (1 - \mathbf{M}) \\
    \mathbf{F}_{\mathrm{final}}^{l,\mathrm{tgt}} &= \alpha \circ \mathbf{F}_{\mathrm{masked}}^{l,\mathrm{tgt}} + (1 - \alpha) \circ \mathbf{F}_{\mathrm{new}}^{l,\mathrm{tgt}}\,,
\end{align}
where $\alpha \in [0, 1]$ is a blending coefficient controlling the influence of the warped attention $\mathbf{F}_{\mathrm{warp}}^{l,\mathrm{tgt}}$, and $\mathbf{M}$ is the binary mask defined in Eq.~\ref{eq:warp_valid_mask}, which ensures that warped attention is only applied to visible regions, while non-visible regions rely solely on the new attention from the target view edit. This blending leads to attention that is correctly applied based on the visibility of regions, maintaining the integrity and realism of the 3D scene during edits. 
\noindent{\textbf{Decaying the Blending Coefficient}: We gradually decay the blend coefficient $\alpha$ during the denoising process to balance the influence of the warped and new attention feature maps and reduce the risk of introducing out-of-distribution features as the process progresses.
 $\alpha_{t} = \alpha_0 \cdot \left(\frac{T - t}{T}\right)$, 
where $\alpha_0=0.9$ is the initial blend coefficient, $t$ is the current denoising timestep, and $T$ is the total number of timesteps. The warped attention feature maps help define the overall structure early in the process, while later iterations focus on refining details, balancing between warped and new maps.

\noindent{\bf Optional Masking with Language-SAM\quad} 
To enhance edit precision, and following previous contributions~\cite{chen2024gaussianeditor, chen2024dge, gaussctrl2024}, we optionally apply a language-guided segmentation mask using Language SAM (combining SAM with Grounding DINO)~\cite{2024slangsam, kirillov2023sam, liu2023groundingDINO}. This mask $\mathbf{M}_{s}$ restricts the diffusion editing process to specific regions of the source view:

\begin{equation}
    \mathbf{I}'_{\mathrm{src}} = \mathrm{DM}\left(\mathbf{I}_{\mathrm{src}}, \mathbf{t}\right) \circ \mathbf{M}_{s}
\end{equation}

The mask ensures that only the targeted regions are modified, preserving the integrity of the surrounding scene. {This approach is particularly effective for edits focusing on specific objects within the image, allowing for precise modifications while leaving the rest of the scene unchanged.}

\noindent{\bf Iterative Optimization\quad} To achieve high consistency and convergence, the editing process is performed iteratively over a small number of iterations (3 iterations in all experiments).  The editing pipeline is summarized in Alg.~\ref{alg:one}; Each iteration involves the following steps:
\begin{enumerate}
    \item \textbf{Subset Editing}: Select a subset of views from the dataset and apply the diffusion-based editing process, generating warped attention feature maps for these views.
    \item \textbf{GS Optimization}: Fine-tune the Gaussian splatting representation $\mathbf{S}$ based on the edited images to align with the modifications.
\end{enumerate}

This iterative approach allows the model to progressively refine the scene, ensuring that edits remain consistent across all views and that the Gaussian splatting representation accurately reflects the desired changes.

\noindent\textbf{Subset Editing}: We follow common practice by working on a subset of the data, using 40 samples to ensure a balance between computational efficiency and comprehensive evaluation. The subset of views is selected randomly from images that have not been edited up to that stage, ensuring diverse perspectives are incorporated without reusing previously edited views. This approach maintains variety and helps capture more comprehensive updates across the scene.
\noindent\textbf{GS Optimization}: Our optimization follows 2DGS~\cite{Huang2DGS2024} and includes L1 and LPIPS~\cite{zhang2018perceptual} RGB losses to measure the discrepancy between the rendered and edited images $\mathbf{I}'$. Additionally, we incorporate normal consistency and depth distortion losses for surface alignment and controlled weight distribution along the rays. \noindent\textbf{L1 Loss}: Calculates pixel-wise absolute differences between edited and target images to maintain fidelity. \noindent\textbf{LPIPS Loss}: Evaluates perceptual similarity using deep feature representations to ensure realistic results. \noindent\textbf{Normal Consistency Loss}: Aligns splat normals with depth map gradients:
 $   \mathcal{L}_n = \sum_{i} \omega_i (1 - n_i^{\mathrm{T}}N)$,
where $i$ indexes intersected splats, $\omega_i$ is the blending weight, $n_i$ is the splat normal facing the camera, and $N$ is the normal from the depth map gradient. \noindent\textbf{Depth Distortion Loss}: Minimizes distance between ray-splat intersections for concentrated weight distribution
 $   \mathcal{L}_d = \sum_{i, j} \omega_i \omega_j |z_i - z_j|$, 
where $\omega_i$ is the blending weight, and $z_i$ is the depth value at the intersection point $i$.

\begin{algorithm}[t]
\caption{Diffusion-Based Attention Warping for 3D Scene Editing\label{alg:one}}
\begin{algorithmic}[1]
\Require Pretrained Gaussian splatting model $\mathbf{S}$, textual instruction $\mathbf{t}$, image $\mathbf{I}$ and reference image or depth map $\mathbf{x}_{\mathrm{tgt}}$, number of stages $S$
\State Generate edited image $\mathbf{I}'_{\mathrm{src}} = \mathrm{DM}(\mathbf{I}, \mathbf{x}_{\mathrm{tgt}}, \mathbf{t})$
\For{stage $s = 1$ to $S$}
    \State Select subset of views $\{\mathbf{I}_{\mathrm{tgt}}\}$
    \For{each target view $\mathbf{I}_{\mathrm{tgt}}$}
        \State Warp $\mathbf{F}_{\mathrm{warp}}^{\mathrm{src}}$ attention feature maps: $\mathbf{F}_{\mathrm{warp}}^{\mathrm{tgt}}$
        \State Generate target view $\mathrm{DM}\left(\mathbf{I}_{\mathrm{tgt}}, {\mathbf{t}}, \mathbf{x}_{\mathrm{tgt}}, \mathbf{F}_{\mathrm{warp}}^{\mathrm{tgt}}\right)$
        \State Fine-tune Gaussian splatting model $\mathbf{S}$ with $\mathbf{I}'_{\mathrm{tgt}}$
    \EndFor
\EndFor
\end{algorithmic}
\end{algorithm}

\renewcommand{\arraystretch}{1.5} 
\setlength{\tabcolsep}{1pt}
\begin{figure*}[ht]
    \centering
    \begin{adjustbox}{max width=\textwidth}
    \begin{tabular}{p{0.15\textwidth}p{0.17\textwidth}p{0.17\textwidth}p{0.17\textwidth}p{0.17\textwidth}p{0.17\textwidth}p{0.17\textwidth}p{0.17\textwidth}}
    \multicolumn{1}{c}{\textbf{Source}} & \multicolumn{1}{c}{\textbf{IGS2GS}} & \multicolumn{1}{c}{\textbf{DGE}} & \multicolumn{1}{c}{\textbf{Ours (IP2P)}} & \multicolumn{1}{c}{\textbf{GC}} & \multicolumn{1}{c}{\textbf{Ours (CN)}} & \multicolumn{1}{c}{\textbf{GC (Random)}} & \multicolumn{1}{c}{\textbf{Ours CN Random}} \\

    \begin{minipage}[t][0.25\textheight][t]{\linewidth} 
        \centering
        \includegraphics[width=0.98\linewidth]{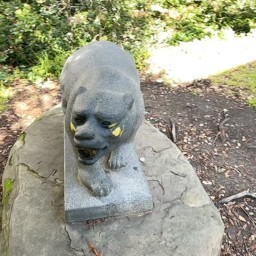}
        \rotatebox{0}{\parbox{0.10\textheight}{~\\\textbf{A bear with rainbow color}}}
    \end{minipage} &
    \begin{minipage}[t][0.22\textheight][t]{\linewidth}
        \centering
        \includegraphics[width=0.98\linewidth]{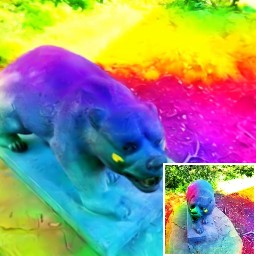} \\
        \vspace{3pt}
        \includegraphics[width=0.98\linewidth]{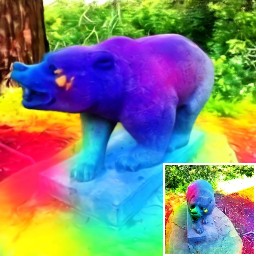}
    \end{minipage} &
    \begin{minipage}[t][0.22\textheight][t]{\linewidth}
        \centering
        \includegraphics[width=0.98\linewidth]{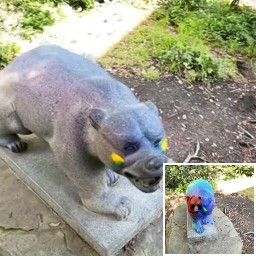} \\
        \vspace{3pt}
        \includegraphics[width=0.98\linewidth]{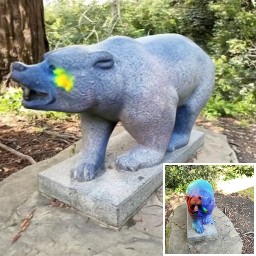}
    \end{minipage} &
    \begin{minipage}[t][0.22\textheight][t]{\linewidth}
        \centering
        \includegraphics[width=0.98\linewidth]{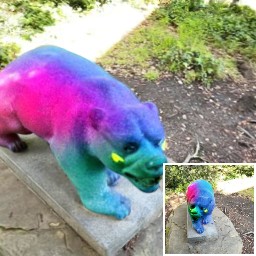} \\
        \vspace{3pt}
        \includegraphics[width=0.98\linewidth]{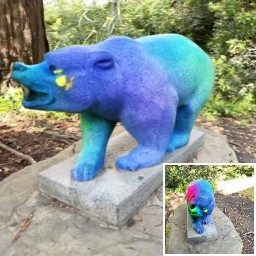}
    \end{minipage} &
    \begin{minipage}[t][0.22\textheight][t]{\linewidth}
        \centering
        \includegraphics[width=0.98\linewidth]{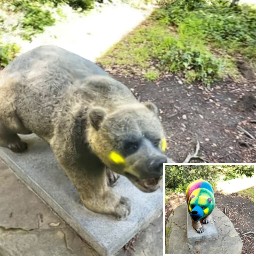} \\
        \vspace{3pt}
        \includegraphics[width=0.98\linewidth]{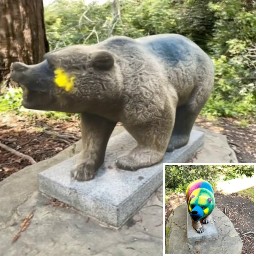}
    \end{minipage} &
    \begin{minipage}[t][0.22\textheight][t]{\linewidth}
        \centering
        \includegraphics[width=0.98\linewidth]{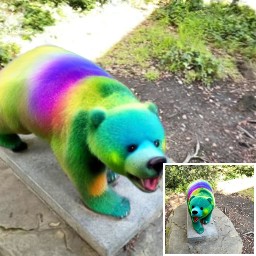} \\
        \vspace{3pt}
        \includegraphics[width=0.98\linewidth]{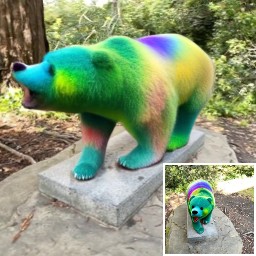}
    \end{minipage} &
    \begin{minipage}[t][0.22\textheight][t]{\linewidth}
        \centering
        \includegraphics[width=0.98\linewidth]{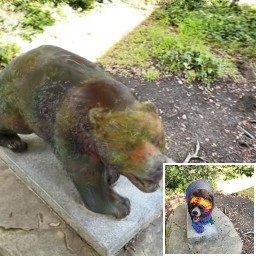} \\
        \vspace{3pt}
        \includegraphics[width=0.98\linewidth]{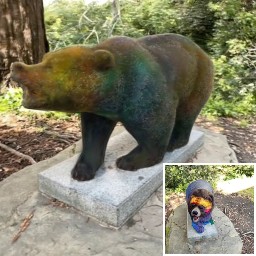}
    \end{minipage} &
    \begin{minipage}[t][0.22\textheight][t]{\linewidth}
        \centering
        \includegraphics[width=0.98\linewidth]{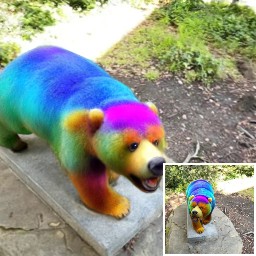} \\
        \vspace{3pt}
        \includegraphics[width=0.98\linewidth]{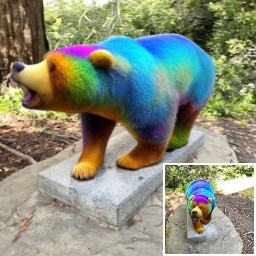}
    \end{minipage} \\

    \begin{minipage}[t][0.25\textheight][t]{\linewidth} 
        \centering
        \includegraphics[width=0.98\linewidth]{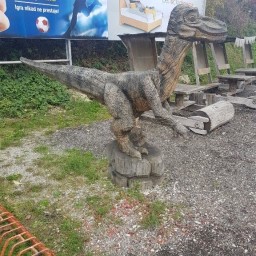}
        \rotatebox{0}{\parbox{0.10\textheight}{~\\ \textbf{A dinosaur statue under water}}}
    \end{minipage} &
    \begin{minipage}[t][0.22\textheight][t]{\linewidth}
        \centering
        \includegraphics[width=0.98\linewidth]{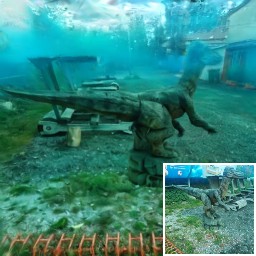} \\
        \vspace{3pt}
        \includegraphics[width=0.98\linewidth]{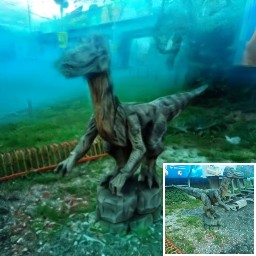}
    \end{minipage} &
    \begin{minipage}[t][0.22\textheight][t]{\linewidth}
        \centering
        \includegraphics[width=0.98\linewidth]{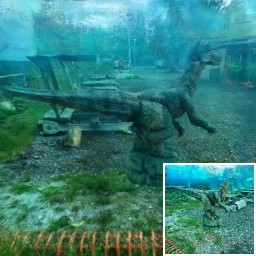} \\
        \vspace{3pt}
        \includegraphics[width=0.98\linewidth]{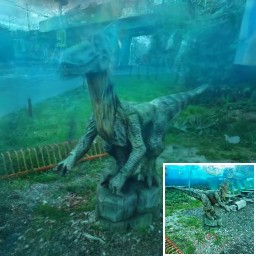}
    \end{minipage} &
    \begin{minipage}[t][0.22\textheight][t]{\linewidth}
        \centering
        \includegraphics[width=0.98\linewidth]{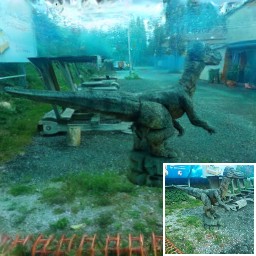} \\
        \vspace{3pt}
        \includegraphics[width=0.98\linewidth]{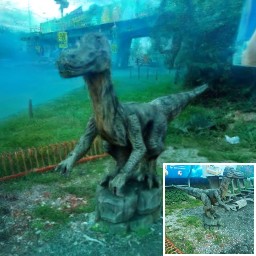}
    \end{minipage} &
    \begin{minipage}[t][0.22\textheight][t]{\linewidth}
        \centering
        \includegraphics[width=0.98\linewidth]{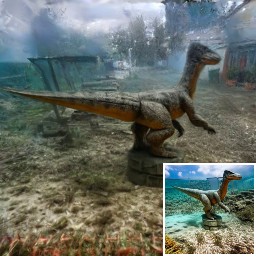} \\
        \vspace{3pt}
        \includegraphics[width=0.98\linewidth]{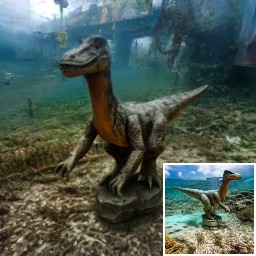}
    \end{minipage} &
    \begin{minipage}[t][0.22\textheight][t]{\linewidth}
        \centering
        \includegraphics[width=0.98\linewidth]{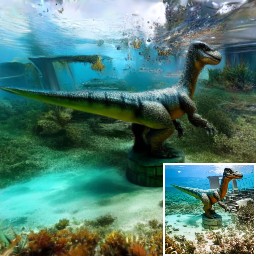} \\
        \vspace{3pt}
        \includegraphics[width=0.98\linewidth]{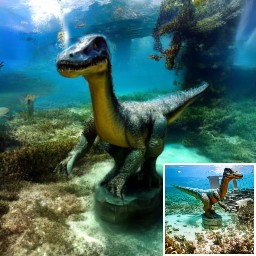}
    \end{minipage} &
    \begin{minipage}[t][0.22\textheight][t]{\linewidth}
        \centering
        \includegraphics[width=0.98\linewidth]{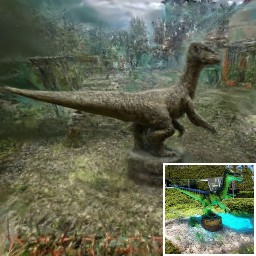} \\
        \vspace{3pt}
        \includegraphics[width=0.98\linewidth]{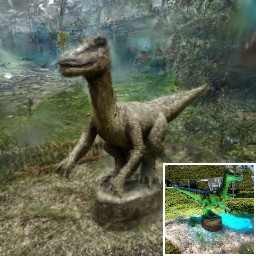}
    \end{minipage} &
    \begin{minipage}[t][0.22\textheight][t]{\linewidth}
        \centering
        \includegraphics[width=0.98\linewidth]{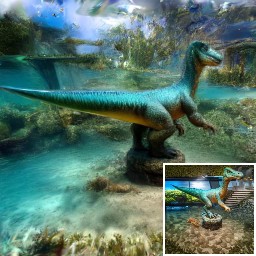} \\
        \vspace{3pt}
        \includegraphics[width=0.98\linewidth]{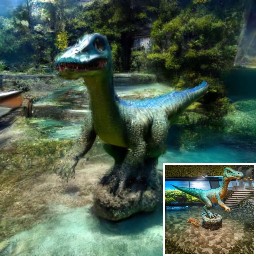}
    \end{minipage} \\

    \begin{minipage}[t][0.25\textheight][t]{\linewidth} 
        \centering
        \includegraphics[width=0.98\linewidth]{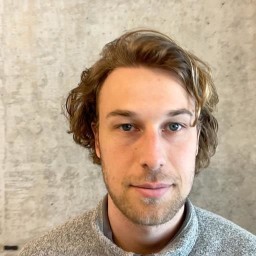}
        \rotatebox{0}{\parbox{0.10\textheight}{~\\ \textbf{A spider man with a mask and curly hair}}}
    \end{minipage} &
    \begin{minipage}[t][0.22\textheight][t]{\linewidth}
        \centering
        \includegraphics[width=0.98\linewidth]{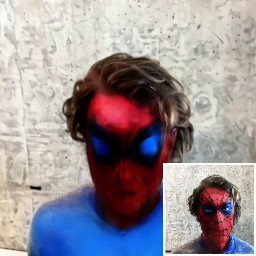} \\
        \vspace{3pt}
        \includegraphics[width=0.98\linewidth]{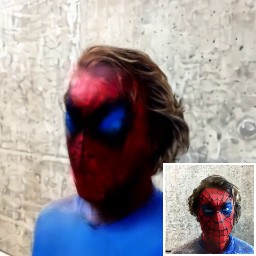}
    \end{minipage} &
    \begin{minipage}[t][0.22\textheight][t]{\linewidth}
        \centering
        \includegraphics[width=0.98\linewidth]{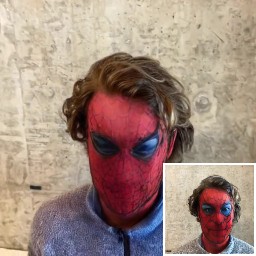} \\
        \vspace{3pt}
        \includegraphics[width=0.98\linewidth]{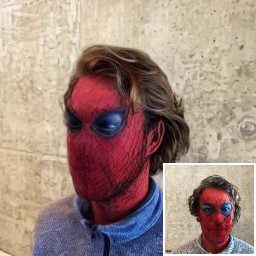}
    \end{minipage} &
    \begin{minipage}[t][0.22\textheight][t]{\linewidth}
        \centering
        \includegraphics[width=0.98\linewidth]{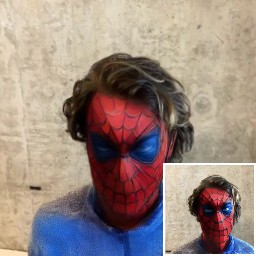} \\
        \vspace{3pt}
        \includegraphics[width=0.98\linewidth]{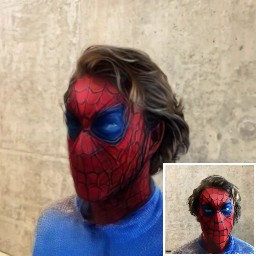}
    \end{minipage} &
    \begin{minipage}[t][0.22\textheight][t]{\linewidth}
        \centering
        \includegraphics[width=0.98\linewidth]{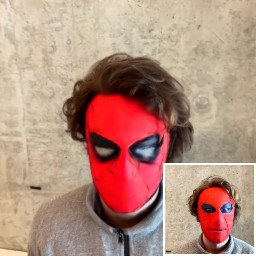} \\
        \vspace{3pt}
        \includegraphics[width=0.98\linewidth]{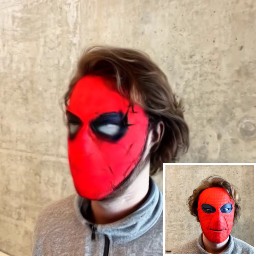}
    \end{minipage} &
    \begin{minipage}[t][0.22\textheight][t]{\linewidth}
        \centering
        \includegraphics[width=0.98\linewidth]{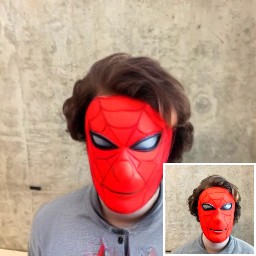} \\
        \vspace{3pt}
        \includegraphics[width=0.98\linewidth]{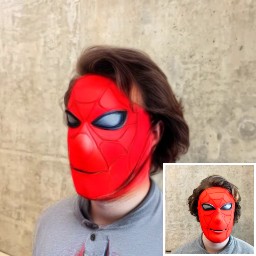}
    \end{minipage} &
    \begin{minipage}[t][0.22\textheight][t]{\linewidth}
        \centering
        \includegraphics[width=0.98\linewidth]{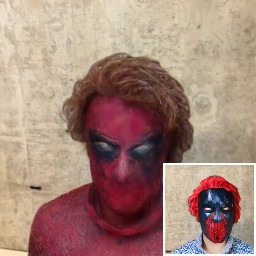} \\
        \vspace{3pt}
        \includegraphics[width=0.98\linewidth]{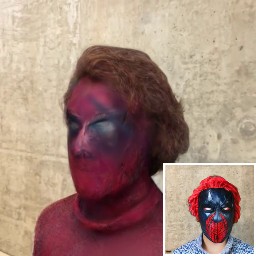}
    \end{minipage} &
    \begin{minipage}[t][0.22\textheight][t]{\linewidth}
        \centering
        \includegraphics[width=0.98\linewidth]{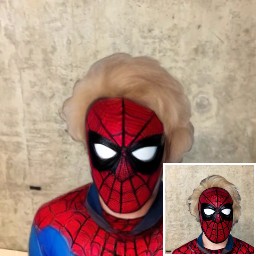} \\
        \vspace{3pt}
        \includegraphics[width=0.98\linewidth]{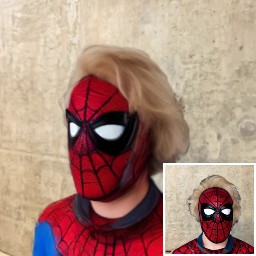}
    \end{minipage} \\

    \begin{minipage}[t][0.25\textheight][t]{\linewidth} 
        \centering
        \includegraphics[width=0.98\linewidth]{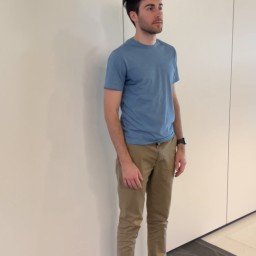}
        \rotatebox{0}{\parbox{0.10\textheight}{~\\ \textbf{A robot stands to a wall}}}
    \end{minipage} &
    \begin{minipage}[t][0.22\textheight][t]{\linewidth}
        \centering
        \includegraphics[width=0.98\linewidth]{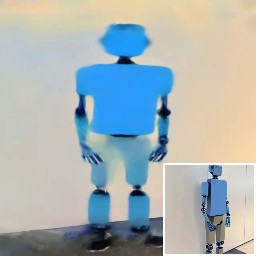} \\
        \vspace{3pt}
        \includegraphics[width=0.98\linewidth]{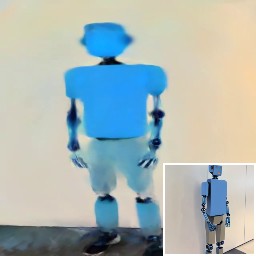}
    \end{minipage} &
    \begin{minipage}[t][0.22\textheight][t]{\linewidth}
        \centering
        \includegraphics[width=0.98\linewidth]{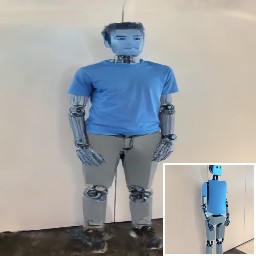} \\
        \vspace{3pt}
        \includegraphics[width=0.98\linewidth]{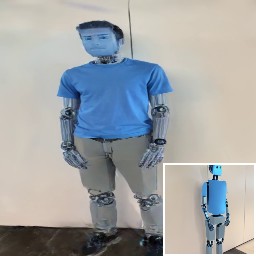}
    \end{minipage} &
    \begin{minipage}[t][0.22\textheight][t]{\linewidth}
        \centering
        \includegraphics[width=0.98\linewidth]{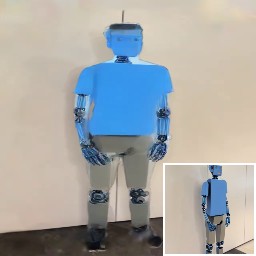} \\
        \vspace{3pt}
        \includegraphics[width=0.98\linewidth]{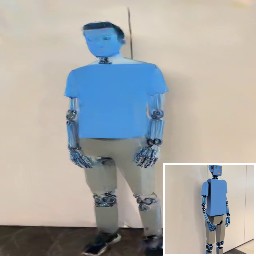}
    \end{minipage} &
    \begin{minipage}[t][0.22\textheight][t]{\linewidth}
        \centering
        \includegraphics[width=0.98\linewidth]{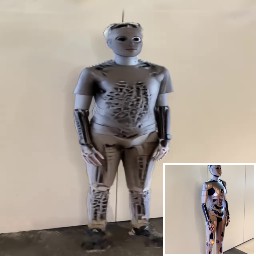} \\
        \vspace{3pt}
        \includegraphics[width=0.98\linewidth]{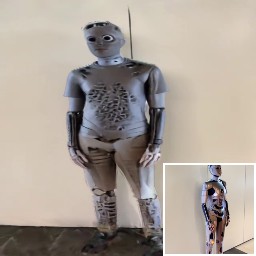}
    \end{minipage} &
    \begin{minipage}[t][0.22\textheight][t]{\linewidth}
        \centering
        \includegraphics[width=0.98\linewidth]{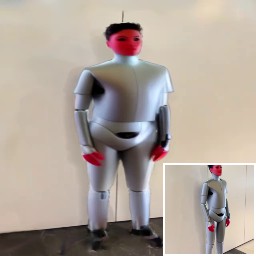} \\
        \vspace{3pt}
        \includegraphics[width=0.98\linewidth]{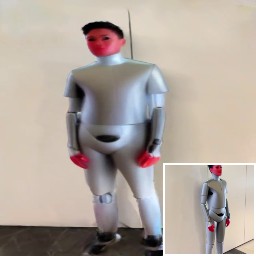}
    \end{minipage} &
    \begin{minipage}[t][0.22\textheight][t]{\linewidth}
        \centering
        \includegraphics[width=0.98\linewidth]{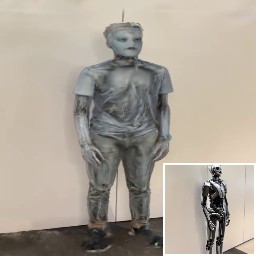} \\
        \vspace{3pt}
        \includegraphics[width=0.98\linewidth]{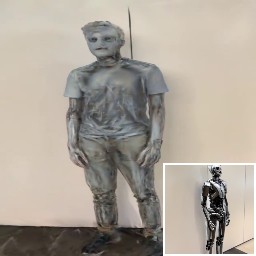}
    \end{minipage} &
    \begin{minipage}[t][0.22\textheight][t]{\linewidth}
        \centering
        \includegraphics[width=0.98\linewidth]{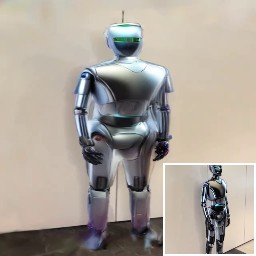} \\
        \vspace{3pt}
        \includegraphics[width=0.98\linewidth]{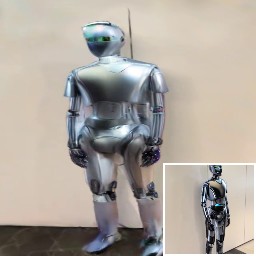}
    \end{minipage} \\

    \end{tabular}
    \end{adjustbox}
    \caption{A comparison of scene editing methods across various scenes is presented. Each sample shows two views, with the modified source image shown as an inset. Additional examples are provided in Figs.~\ref{fig:main_vis_bear}, \ref{fig:main_vis_dino}, \ref{fig:main_vis_face}, \ref{fig:main_vis_person} and \ref{fig:main_vis_table}.}\label{fig:comparisons}

\end{figure*}
\renewcommand{\arraystretch}{1.0} 
\setlength{\tabcolsep}{1pt}

\begin{figure}[!htbp]
    \centering
    \includegraphics[width=\linewidth]{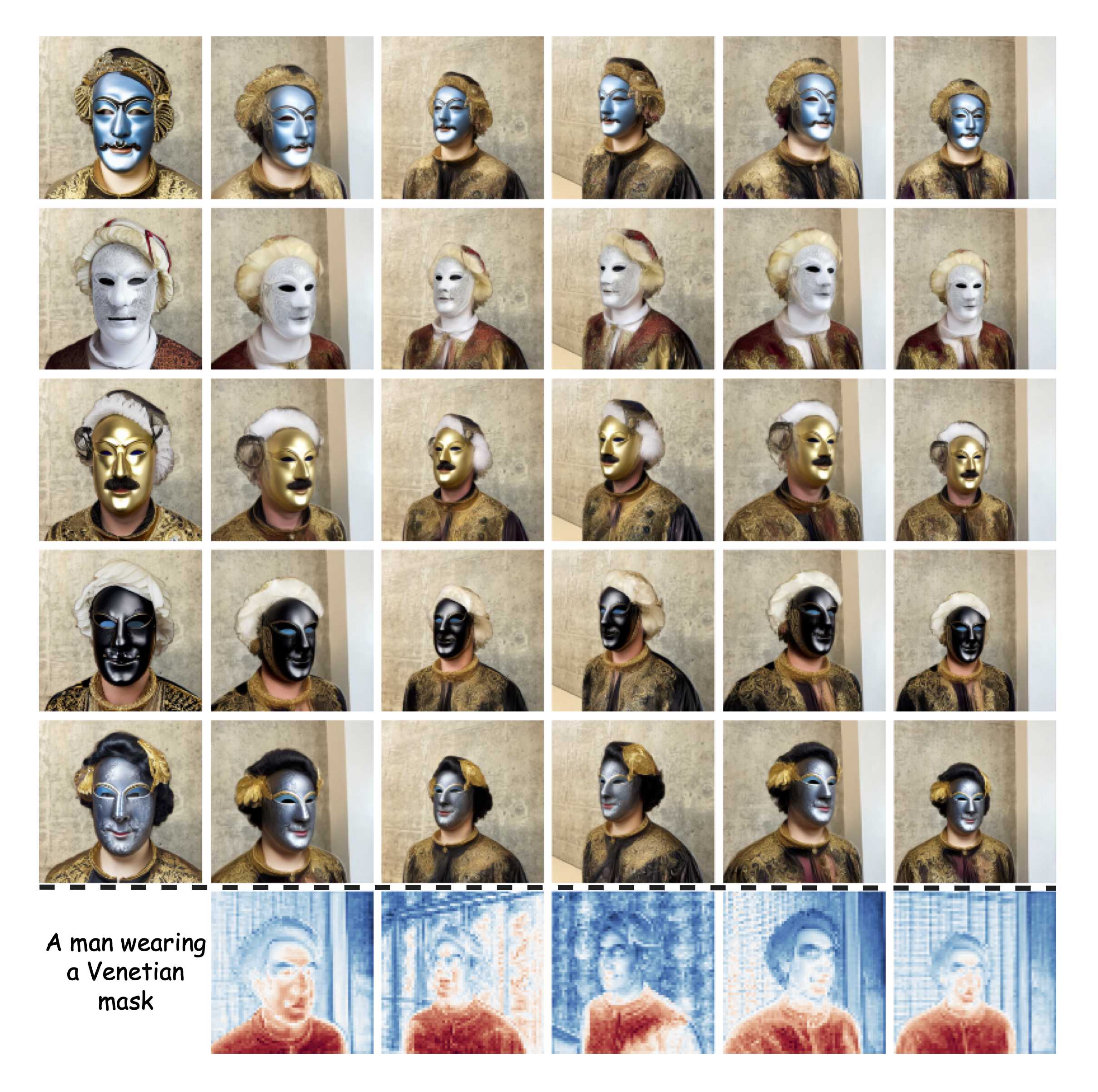}
    \caption{Obtaining variability. Our method (ControlNet variant) with different random seeds to produce diverse stylistic variations. Each row illustrates how varying the random seed impacts the visual output, resulting in unique edits while preserving the overall content structure. Additionally, the figure includes the warped feature map of the source view (left column) to provide insight into how the attention is distributed across the edited images. \vspace{-.5cm}}
    
    \label{fig:vis_seeds_controlnet}
\end{figure}

\section{Experiments}
\label{sec:exp}

\begin{table*}[t]
\centering
\begin{tabular}{@{}l@{~~}c@{~~}c@{~~}c@{~~}c@{~~}c@{}}
\toprule
\textbf{Method} & \textbf{Edit PSNR} & \textbf{Clip Similarity} & \textbf{Clip Dir. Sim.} & \textbf{DINO Single Img Sim.} & \textbf{Clip Single Img Sim.} \\
\midrule
IGS2GS (default params) & 19.745 & 0.253 & 0.150 & 0.518 & 0.771 \\
{IGS2GS (improved)} & 21.429 & 0.267 & 0.149 & 0.627 & 0.841 \\
DGE & 25.667 & 0.257 & 0.160 & 0.692 & 0.862 \\
\textbf{Ours IP2P} & 22.780 & 0.264 & 0.144 & 0.727 & \textbf{0.868} \\
\midrule
GaussCtrl & \textbf{26.575} & 0.252 & 0.138& 0.670 & 0.843  \\
\textbf{Ours ControlNet} & 22.522 & 0.268 & 0.172 & \textbf{0.756} & 0.859 \\
GaussCtrl-Random & 20.438 & 0.236 & 0.103 & 0.517 & 0.767 \\
\textbf{Ours ControlNet Random} & 20.212 & \textbf{0.276} & \textbf{0.189} & 0.743 & 0.852 \\
\bottomrule
\end{tabular}%
\caption{Comparison of Methods: The methods above the separator use InstructPix2Pix for prompt-based editing, while those below rely on ControlNet-based editing.}
\label{tab:metrics_part1}
\end{table*}

To thoroughly evaluate our method, we conducted tests on six diverse scenes using 17 unique prompts to assess performance across a range of scenarios. These included the same scenes used for the evaluation of DGE~\cite{chen2024dge}, enabling direct comparison. The selected prompts covered object-centric and non-object-centric scenes, indoor and outdoor environments, and human face edits. While detailed evaluation data is not consistently provided in prior works, we emphasize transparency to encourage reproducibility and comparability in future research.

We evaluated our method on several benchmark datasets commonly used in 3D scene editing and rendering tasks: IN2N~\cite{instructnerf2023}, Mip-NeRF360~\cite{barron2022mipnerf360}, and BlendedMVS~\cite{yao2020blendedmvs}. All experiments were conducted on a 512×512 images. These datasets challenge our method with varied lighting, complex geometries, and textures, demonstrating its adaptability across scenarios.

\noindent{\bf Baselines\quad}
We compared our method with several recent state-of-the-art baselines, including IGS2GS~\cite{igs2gs} (a Gaussian splatting version of IN2N~\cite{instructnerf2023}), GaussCtrl~\cite{gaussctrl2024}, and DGE~\cite{chen2024dge}. IGS2GS and DGE are based on the InstructPix2Pix~\cite{brooks2022instructpix2pix} diffusion model for image editing, while GaussCtrl leverages ControlNet~\cite{zhang2023adding}. These baselines were chosen for their demonstrated ability to produce high-quality edits and their relevance to 3D editing, allowing us to comprehensively evaluate how our method performs against the current state-of-the-art. Additional promising methods, such as VCEdit~\cite{wang2025view} and LatentEditor~\cite{khalid2023latenteditor}, were not included in our experiments due to the lack of publicly available implementations.

\noindent{\bf Evaluation Metrics\quad}
To provide a comprehensive evaluation of our method, we employed a range of metrics covering both standard and perceptual measures. While prior works often focus on a limited set of metrics, we aimed for a broader assessment to give a complete overview of our method's performance. Next, we outline the metrics used. \noindent\textbf{Edit PSNR}: This metric calculates the Peak Signal-to-Noise Ratio (PSNR) between the edited images generated by the diffusion model and the rendered images, quantifying the fidelity of the edits. \noindent\textbf{CLIP Similarity}: A standard metric for perceptually comparing images and text. We encode the training set using the CLIP~\cite{radford2021learningtransferablevisualmodels} model and separately encode the target prompt into CLIP space. The cosine similarity between these encodings measures how closely the edited images align with the intended target prompt. \noindent\textbf{CLIP Directional Similarity}: This metric~\cite{gal2022stylegan} assesses the consistency of changes between images and text. We compute the cosine similarity between:
1. The difference in CLIP space between the original training set and the rendered training set.
2. The difference between the source and target prompts.
This metric captures how well the direction of change in the image corresponds to the intended change described by the prompts. 

We also included two custom metrics to evaluate how well the rendered training set matches the edited image: \noindent\textbf{DINO Single Image Similarity}: This metric calculates the mean similarity in DINOv2~\cite{oquab2024dinov2learningrobustvisual, caron2021emerging} space between the edited source image and the training set. DINO embeddings are known for capturing detailed visual semantics, making this metric effective for assessing visual alignment with the target appearance. \noindent\textbf{CLIP Single Image Similarity}: Similar to DINO Single Image Similarity, but with CLIP embeddings.

\noindent\textbf{Results\quad}
We evaluated our method against state-of-the-art techniques in two main categories: models based on InstructPix2Pix (IP2P), including IGS2GS and DGE, and models based on ControlNet guided by depth (GuassCtrl). GaussCtrl was tested using both latent inversion and random latent initialization methods. In the table, the label “Random” indicates the use of ControlNet with a randomly initialized latent, as opposed to the latent inversion approach~\cite{song2020denoising} employed in GaussCtrl. IGS2GS is applied both with its default parameters and with a better set of parameters we found, in which the main difference is using fewer iterations (3000 instead of 7000).

As shown in Table~\ref{tab:metrics_part1}, our method outperforms all compared models in single-image similarity metrics, demonstrating superior alignment with the target images in both DINO and CLIP spaces. For CLIP similarity and directional similarity, our method shows better performance than GaussCtrl. In comparison with IP2P-based models, while there are instances where our method shows minor differences, it often performs on par or better. Recognizing that these metrics may not fully capture the subjective quality of edits, we also conducted qualitative analyses and user studies for a more comprehensive evaluation. Although some methods achieve higher scores in the Edit PSNR metric, it is important to note that this metric is biased, since it favors methods that edit fewer reference images as part of their 3D model finetuning. When fewer images are edited, the 3D resulting model is better fitted to these views, especially since the information is propagated in such a way that the image editing is based on the progress of the 3D model. This, of course, comes at the price of relying on reference views that are inconsistent with the prompt. 

 To complement our quantitative evaluation, 
Fig.~\ref{fig:comparisons} compares our method against the baselines across various scenes and prompts, showcasing how our approach maintains superior edit quality and consistency. Note that significant differences in the edited reference images between methods arise from the way consistency is achieved between the reference views, even for the same diffusion model seed.

Lastly, Fig.~\ref{fig:vis_seeds_controlnet} demonstrates the versatility of our single-image editing capability, where we can apply edits using the same prompt but achieve different styles. The GaussCtrl method shows no versatility due to the reliance on DDIM inversion~\cite{zhang2023inversion}. The other baselines  obtain different results for different seeds, but due to their propagation mechanism, display less variability, see Fig.~\ref{fig:seed_comparisons}. 

\noindent\textbf{User Study\quad}
We conducted a user study involving 28 participants, each answering 10 questions across 6 different scenes. In each question, users were shown a source image, a prompt, and two novel views generated by each method. Participants were asked to choose the image they believed best represented the edit, focusing on both edit quality and coherence with the given prompt. We did not limit participants' evaluation criteria, allowing them to assess based on their judgment. To ensure a fair comparison, the questions were divided into two sets for IP2P-based models and for ControlNet-based models, preventing cross-model biases during evaluation. The results of the user study are summarized in Table~\ref{tab:user_study_results}. Evidently, there is a clear reference to our method in both groups, where in the second there is an advantage to the method that is randomly initialized.

\begin{table}[t]
    \centering
    \begin{tabular}{l@{~~~~~~}c}
        \toprule
        \textbf{Method} & \textbf{Selection Frequency} \\
        \midrule
        DGE & 0.142 \\
        IGS2GS & 0.265 \\
        Ours IP2P & \textbf{0.594} \\
        \midrule 
        GaussCtrl & 0.148 \\
        GaussCtrl Random & 0.058 \\
        Ours ControlNet & 0.154 \\
        Our ControlNet Random & \textbf{0.639} \\
        \bottomrule
    \end{tabular}
        \caption{User study results for the two groups of methods.}
    \label{tab:user_study_results}
\end{table}

\begin{table}[ht]
    \centering
    \resizebox{\columnwidth}{!}{%
    \begin{tabular}{@{}l@{~}c@{~}c@{~}c@{~}c@{~}c@{}}
        \hline
        \textbf{Name} & \textbf{PSNR} & \textbf{C. Sim.} & \textbf{Dir. Sim.} & \textbf{D. Im. Sim.} & \textbf{C. Im. Sim.} \\
        \hline
        1 stage & 18.015 & 0.271 & 0.175 & 0.726 & 0.826 \\
        2 stage & 17.196 & 0.278 & 0.205 & 0.768 & 0.871 \\
        Only SA & 18.992 & 0.278 & 0.209 & 0.771 & \textbf{0.879} \\
        W/O decay & 20.123 & 0.277 & 0.210 & 0.767 & 0.871 \\
        W/O $M$ & \textbf{20.661} & 0.280 & 0.203 & 0.765 & 0.873 \\
        Full method & 18.949 & \textbf{0.281} & \textbf{0.211} & \textbf{0.773} & 0.877 \\
        \hline
    \end{tabular}%
    }
        \caption{Ablation study results. C=Clip D=Dino.}
    \label{tab:ablation_results}
\end{table}

\noindent{\bf Ablation Study\quad} We conducted thorough ablations to evaluate the impact of different components of our method, including: 
\textbf{Warping mask}: removing the warping mask as described in Equation~\ref{eq:warp}. \textbf{Blending decay mechanism}: a constant blending coefficient throughout the denoising process, rather than decaying it over time. \textbf{Self-attention}: injecting only self-attention feature maps into the diffusion model. \textbf{Iterative optimization}: We ran our method using both single-stage and two-stage executions instead of the three iterations we use in our experiments.  The ablations were performed on four different scenes from different datasets. 
The results of these ablations are summarized in Table~\ref{tab:ablation_results}. As can be seen, each component contributes to the overall success. The number of iterations helps in an incremental way, mostly to the Clip Directional Similarity score. Removing the cross attention from the blending deteriorates the metrics that measure text alignment, but improves the clip image similarity. The decay and the mask seem to assist in multiple metrics. Various ablations improve the PSNR. However, as mentioned above, this metric can be high despite the overall results being weaker. A visual comparison can be found in Fig.~\ref{fig:psnr}.

\section{Discussion and Limitations}
\label{sec:discussion}

The success of our attention-warping approach in maintaining consistency across multiple views suggests broader applications beyond static 3D scenes. A natural extension would be to apply similar principles to video editing, where optical flow could replace depth-based warping for propagating edits across frames. Unlike recent video editing methods like~\cite{Wu_2023_ICCV,khachatryan2023text2video,geyer2023tokenflow} that rely on temporal diffusion models or frame-by-frame processing, our attention-warping technique could potentially offer more precise control over edit propagation while maintaining temporal coherence, without the necessity of multiple frames processing. This approach would be particularly advantageous compared to methods that depend on direct feature matching or temporal consistency losses, as our warped based attention approach could better preserve fine-grained edit details while ensuring smooth transitions between frames.
The success of our approach in handling occlusions and view-dependent artifacts through the visibility mask and the gaussians normal mask also provides insights into how attention mechanisms can be made more geometry-aware, which could be valuable for improving other 3D-aware generation and editing methods.

However, our method does face several important limitations.
{\textbf{Geometry Dependence:}} The quality of our edits heavily relies on the accuracy of the underlying geometric reconstruction. In cases where the Gaussian splatting model fails to capture accurate depth information or produces noisy geometry, the warping process can lead to artifacts or inconsistent edits across views. {\textbf{Limited Edit Scope:}} While our method handles a wide range of edits, it can struggle with certain types of modifications that require significant geometric changes or involve heavy occlusions. For example, adding large objects that should appear consistently across multiple views remains challenging, as the method primarily focuses on appearance changes rather than structural modifications. {\textbf{Diffusion Model Constraints:}} Our method, which is based on diffusion models, can be limited by their inherent capabilities and limitations.

\section{Conclusions}
We have presented a novel approach for consistent 3D scene editing that leverages attention features from diffusion models through geometric-aware warping. By capturing edit intentions from a single reference view and systematically propagating them across multiple viewpoints, our method achieves high-quality, consistent edits while avoiding the computational overhead of processing multiple views simultaneously. Extensive experiments demonstrate that our approach outperforms existing methods in maintaining edit fidelity across viewpoints, as validated through both quantitative metrics and user studies.

{
    \small
    \bibliographystyle{ieeenat_fullname}
    \bibliography{main}
}

\clearpage
\newpage

\appendix

\renewcommand{\topfraction}{0.9}
\renewcommand{\bottomfraction}{0.8}
\renewcommand{\textfraction}{0.05}
\renewcommand{\floatpagefraction}{0.99}
\renewcommand{\thefigure}{\Roman{figure}}
\renewcommand{\thetable}{\Roman{table}}


\section{Advantages of Single View Editing}
In this section, we present a comprehensive discussion on the benefits of single-view editing in 3D scene editing. Although certain methods can achieve impressive edits with multi-view consistency~\cite{chen2024dge, gaussctrl2024, wang2025view, khalid2023latenteditor}, they lack the flexibility to allow users to control the desired edit style. Why is this control important? It enables users to customize scene edits precisely to their preferences. Additionally, it provides the option to edit a single image without involving a diffusion model by using a single inversion process~\cite{zhang2023inversion} to propagate edits across the scene.

Figure~\ref{fig:seed_comparisons} illustrates visual examples of our results, showing edits made with the same source image and prompt but varying in style. We also provide comparisons to other methods. It is important to note that methods like GaussCtrl~\cite{gaussctrl2024} and Ours ControlNet, which rely on image inversion, are excluded from these comparisons as they use pre-defined inverted images rather than random latents.

\textbf{Limitations of Other Methods}: While some existing methods, such as GaussCtrl, do not offer the flexibility to choose a random edit style due to their reliance on image inversion, we adapted GaussCtrl to support random-based generation for comparison purposes. However, even with this adaptation, GaussCtrl struggles to align consistently with a single source edit style, as demonstrated in Figure~\ref{fig:seed_comparisons}. Although methods like IGS2GS~\cite{igs2gs} and DGE~\cite{chen2024dge} can generate edits using different seeds, they have significant limitations. IGS2GS is capable of producing varying results; however, the generated edits are often of lower quality and lack the ability to allow users to select a specific style explicitly. On the other hand, DGE struggles to produce noticeable variations in edits, even when using different seeds. 
As shown in Figure~\ref{fig:seed_comparisons}, these methods may produce edits that are inconsistent with the initial source-edited image or fail to offer meaningful diversity. For instance, as shown in the first example, although the source image has pink skin, both IGS2GS and DGE generate results with white skin. In contrast, our method is not constrained by the chosen edit style. Users can freely select any edited image to use as the basis for their scene edits, providing superior flexibility and user control over the editing process.
\textbf{User-Generated Edits:} As further illustrated in Figure~\ref{fig:ddpm_inversion}, we provide a visualization of the DDPM inversion~\cite{huberman2024edit} process. In this setup, the user supplies a non-diffusion-based edited image, which is inverted into the latent space using DDPM inversion. Our method is then applied to this inverted edit. The top row in the figure shows the user-generated edited image followed by three novel views generated using our method, while the bottom row displays the user-generated edited image and three corresponding views generated using the DGE~\cite{chen2024dge} method. This comparison demonstrates that, although DGE is a state-of-the-art view-consistent editing method, it requires more than a single input image to produce such edits. In contrast, our method achieves high-quality, consistent results directly from a single user-provided edited image, highlighting a key advantage of our approach.

\section{Ablation Study Details}

In this section, we provide a detailed discussion and visual comparison for the ablation study of our method, as shown in Figure~\ref{fig:ablation}. All ablation experiments are conducted using the \textit{"Ours ControlNet Random"} method. Below, we analyze the impact of different components of our approach:

\textbf{Iterative Process}: We examine the effect of using a non-iterative process (\textit{1-stage} in the figure). It is evident that using only a single stage results in the method struggling to generate coarse and fine details.

\textbf{Self-Attention Only}: This ablation demonstrates the effect of injecting only the attention feature maps from the self-attention layers. While this produces reasonable results, it struggles to reconstruct finer details, such as forehead wrinkles.

\textbf{Without Decay}: Here, we evaluate the impact of removing the decay mechanism for alpha blending between the warped and new attention feature maps. As shown in the figure, the absence of this mechanism results in suboptimal blending and reduced image quality.

\textbf{Without Warp Mask}: In this scenario, the warping mask is omitted during the warping process, leading to visible artifacts caused by the introduction of out-of-distribution features during image generation. For instance, noticeable color leakage occurs between visible and occluded regions, such as the eyelid area and the left side of the man's hair, as well as between the lips and the neck.

Each ablation highlights the importance of the respective components in ensuring high-quality, consistent, and artifact-free results. We conduct our ablations on \textit{face}, \textit{bear} and \textit{dinosaur} scenes from the datasets accordingly~\cite{instructnerf2023, barron2022mipnerf360, yao2020blendedmvs}.

\section{Edit PSNR Visual Comparison}
As highlighted in the main paper, the \textit{Edit PSNR} metric is inherently biased and often fails to reflect the true quality of edits accurately. To illustrate this limitation, we present visual examples in Figure~\ref{fig:psnr}. These examples demonstrate that edits with result in poor visual quality can have high PSNR values. For instance, while GaussCtrl~\cite{gaussctrl2024} achieves a significantly higher Edit PSNR score compared to our method, its final edit does not adhere to the given prompt: \textit{"a photo of a rainbow-colored bear in the forest."} This discrepancy underscores the inadequacy of relying solely on PSNR as a metric for evaluating edit quality.

\section{Method Comparison Visualization}

As outlined in the main paper, we provide additional visual examples comparing our method with other approaches. These visualizations are organized by different scenes and are presented in Figures~\ref{fig:main_vis_bear}, \ref{fig:main_vis_dino}, \ref{fig:main_vis_face}, \ref{fig:main_vis_person}, and \ref{fig:main_vis_table}.

\section{Evaluation Setup and Details}
In this section, we outline the scenes and prompts used for evaluating our method. Our evaluation follows the same setup as DGE~\cite{chen2024dge}, applied to three common scenes, along with an additional three scenes from other datasets. The evaluated scenes include: \textit{Face}, \textit{Bear}, and \textit{Person} from IN2N~\cite{instructnerf2023}; \textit{Garden} and \textit{Stump} from MIP-NeRF360~\cite{barron2022mipnerf360}; and \textit{Dinosaur} from BlendedMVS~\cite{yao2020blendedmvs}. 
The editing prompts, tailored for both IP2P and ControlNet~\cite{brooks2022instructpix2pix, zhang2023adding}, as well as the source and target prompts used for metrics evaluation, are detailed in Tab.~\ref{tab:scene_prompts}.

\section{User Study and Evaluation}
As described in the main paper, we conducted a user study to evaluate the subjective quality of the results. To illustrate the structure of the study, we provide an example question used during the evaluation in Figure~\ref{fig:user_study_example}. This example highlights how participants were asked to compare and assess the quality of edits generated by different methods.

\section{2DGS and 3DGS Warping}
In our paper, we chose to use 2DGS~\cite{Huang2DGS2024} over 3DGS~\cite{kerbl3Dgaussians} due to its superior geometric accuracy. Figure~\ref{fig:2d3dgs_warp} shows the source view, the validity mask highlighting reliable depth regions, and the warping results using 3DGS and 2DGS. While 3DGS produces noticeable artifacts, 2DGS demonstrates better alignment and accuracy. This improvement is particularly evident in regions defined as valid by the mask, further supporting the decision to use 2DGS in our method.

\section{Code and Implementation Details}
We extend our gratitude to NeRFStudio~\cite{nerfstudio}, whose infrastructure served as the foundation for our implementation, providing the necessary tools and framework for developing our method. The code specific to our method (i.e., not including the base code of NeRFStudio) is included in this supplementary material. We also acknowledge IGS2GS~\cite{igs2gs} for additional reference.

\clearpage
\renewcommand{\arraystretch}{1.5} 
\setlength{\tabcolsep}{1pt}
\begin{figure*}[ht]
    \centering
    \begin{adjustbox}{max width=\textwidth}
    \begin{tabular}{p{0.13\textwidth}p{0.13\textwidth}p{0.13\textwidth}p{0.13\textwidth}p{0.13\textwidth}}
    \multicolumn{1}{c}{\textbf{IGS2GS}} & \multicolumn{1}{c}{\textbf{DGE}} & \multicolumn{1}{c}{\textbf{Ours (IP2P)}} & \multicolumn{1}{c}{\textbf{GC (Random)}} & \multicolumn{1}{c}{\textbf{Ours CN Random}} \\

    \begin{minipage}[t][0.13\textheight][t]{\linewidth}
        \centering
        \includegraphics[width=0.91\linewidth]{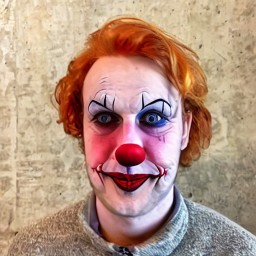} \\
        \includegraphics[width=0.91\linewidth]{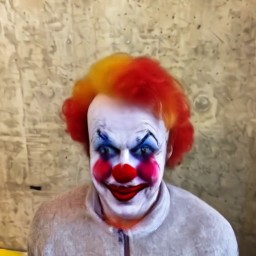} \\
        \includegraphics[width=0.91\linewidth]{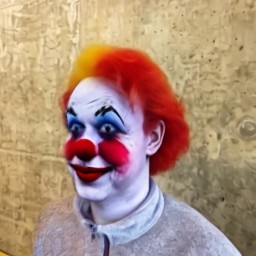}
    \end{minipage} &
    \begin{minipage}[t][0.13\textheight][t]{\linewidth}
        \centering
        \includegraphics[width=0.91\linewidth]{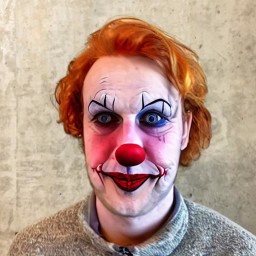} \\
        \includegraphics[width=0.91\linewidth]{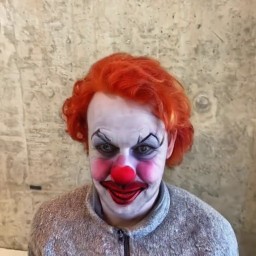} \\
        \includegraphics[width=0.91\linewidth]{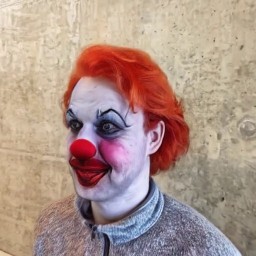}
    \end{minipage} &
    \begin{minipage}[t][0.13\textheight][t]{\linewidth}
        \centering
        \includegraphics[width=0.91\linewidth]{supp_vers_42_ours_ip2p_edit.jpg} \\
        \includegraphics[width=0.91\linewidth]{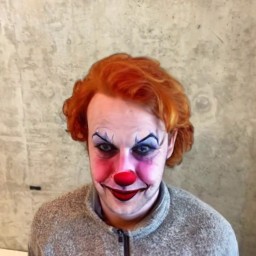} \\
        \includegraphics[width=0.91\linewidth]{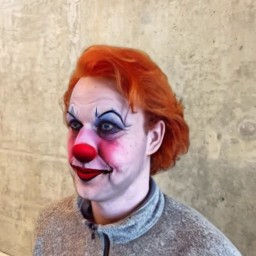}
    \end{minipage} &
    \begin{minipage}[t][0.13\textheight][t]{\linewidth}
        \centering
        \includegraphics[width=0.91\linewidth]{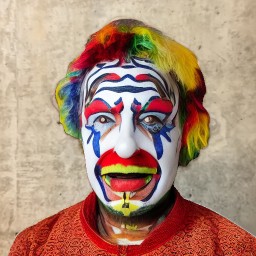} \\
        \includegraphics[width=0.91\linewidth]{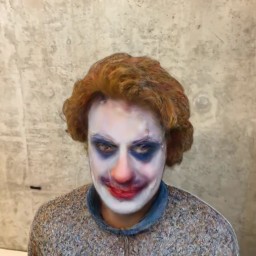} \\
        \includegraphics[width=0.91\linewidth]{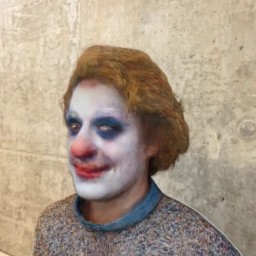}
    \end{minipage} &
    \begin{minipage}[t][0.13\textheight][t]{\linewidth}
        \centering
        \includegraphics[width=0.91\linewidth]{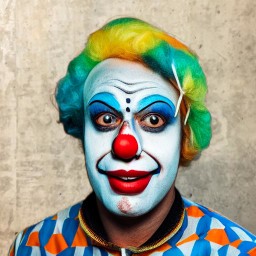} \\
        \includegraphics[width=0.91\linewidth]{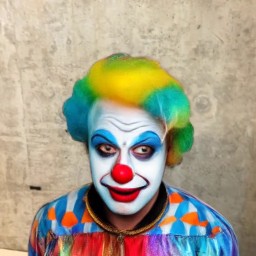} \\
        \includegraphics[width=0.91\linewidth]{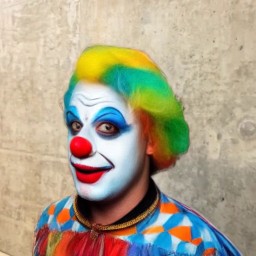}
    \end{minipage} \\[120pt]
\midrule
    \begin{minipage}[t][0.13\textheight][t]{\linewidth}
        \centering
        \includegraphics[width=0.91\linewidth]{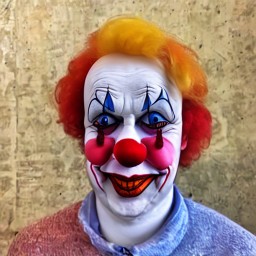} \\
        \includegraphics[width=0.91\linewidth]{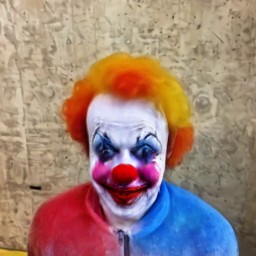} \\
        \includegraphics[width=0.91\linewidth]{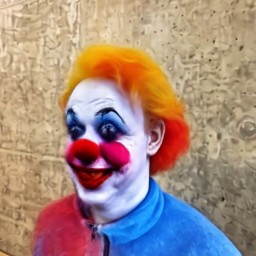}
    \end{minipage} &
    \begin{minipage}[t][0.13\textheight][t]{\linewidth}
        \centering
        \includegraphics[width=0.91\linewidth]{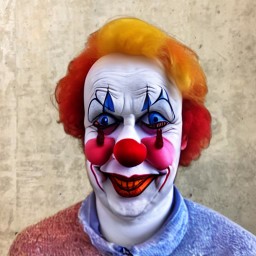} \\
        \includegraphics[width=0.91\linewidth]{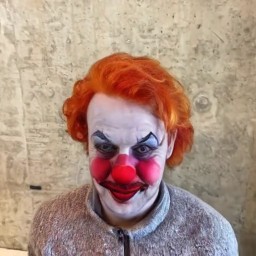} \\
        \includegraphics[width=0.91\linewidth]{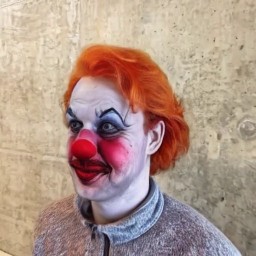}
    \end{minipage} &
    \begin{minipage}[t][0.13\textheight][t]{\linewidth}
        \centering
        \includegraphics[width=0.91\linewidth]{supp_vers_74778_ours_ip2p_edit.jpg} \\
        \includegraphics[width=0.91\linewidth]{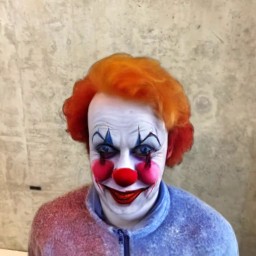} \\
        \includegraphics[width=0.91\linewidth]{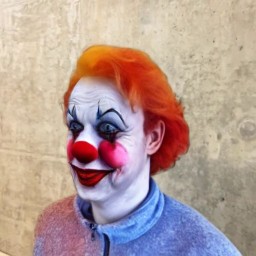}
    \end{minipage} &
    \begin{minipage}[t][0.13\textheight][t]{\linewidth}
        \centering
        \includegraphics[width=0.91\linewidth]{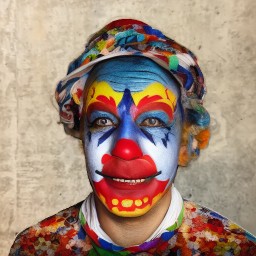} \\
        \includegraphics[width=0.91\linewidth]{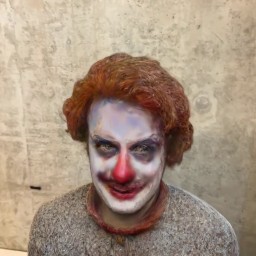} \\
        \includegraphics[width=0.91\linewidth]{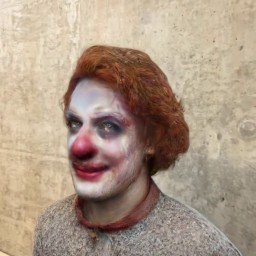}
    \end{minipage} &
    \begin{minipage}[t][0.13\textheight][t]{\linewidth}
        \centering
        \includegraphics[width=0.91\linewidth]{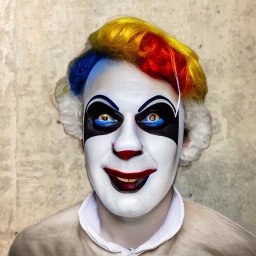} \\
        \includegraphics[width=0.91\linewidth]{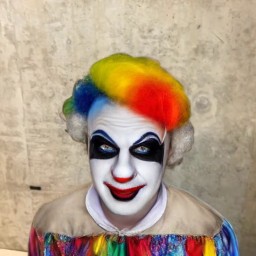} \\
        \includegraphics[width=0.91\linewidth]{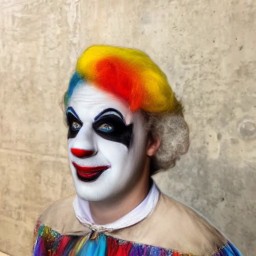}
    \end{minipage} \\[120pt]
\midrule
    \begin{minipage}[t][0.13\textheight][t]{\linewidth}
        \centering
        \includegraphics[width=0.91\linewidth]{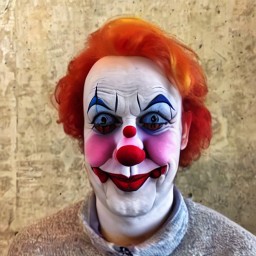} \\
        \includegraphics[width=0.91\linewidth]{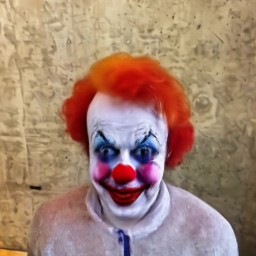} \\
        \includegraphics[width=0.91\linewidth]{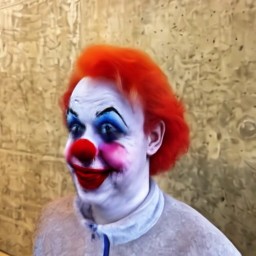}
    \end{minipage} &
    \begin{minipage}[t][0.13\textheight][t]{\linewidth}
        \centering
        \includegraphics[width=0.91\linewidth]{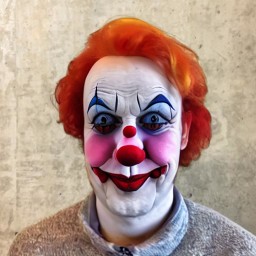} \\
        \includegraphics[width=0.91\linewidth]{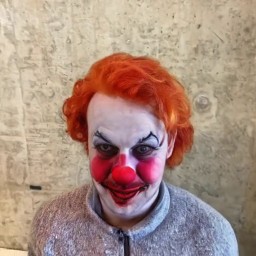} \\
        \includegraphics[width=0.91\linewidth]{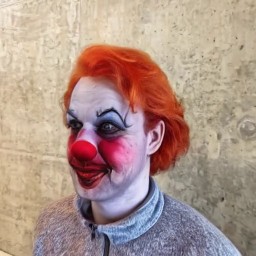}
    \end{minipage} &
    \begin{minipage}[t][0.13\textheight][t]{\linewidth}
        \centering
        \includegraphics[width=0.91\linewidth]{supp_vers_14252_ours_ip2p_edit.jpg} \\
        \includegraphics[width=0.91\linewidth]{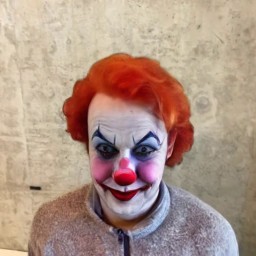} \\
        \includegraphics[width=0.91\linewidth]{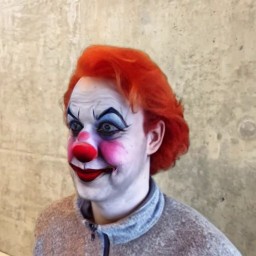}
    \end{minipage} &
    \begin{minipage}[t][0.13\textheight][t]{\linewidth}
        \centering
        \includegraphics[width=0.91\linewidth]{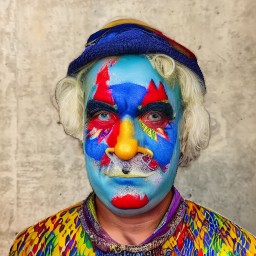} \\
        \includegraphics[width=0.91\linewidth]{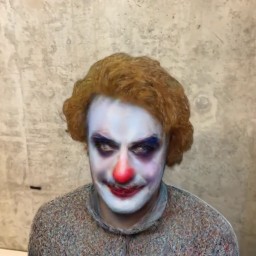} \\
        \includegraphics[width=0.91\linewidth]{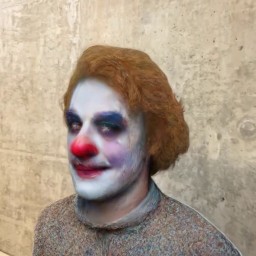}
    \end{minipage} &
    \begin{minipage}[t][0.13\textheight][t]{\linewidth}
        \centering
        \includegraphics[width=0.91\linewidth]{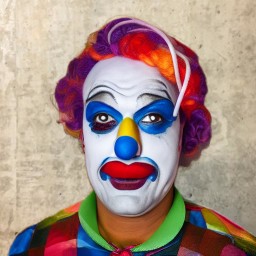} \\
        \includegraphics[width=0.91\linewidth]{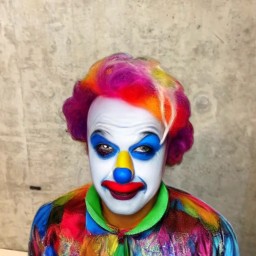} \\
        \includegraphics[width=0.91\linewidth]{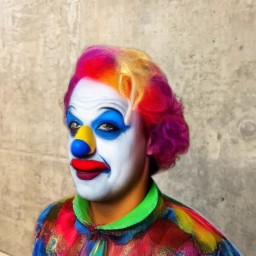}
    \end{minipage} \\
    \end{tabular}
    \end{adjustbox}
    \vspace{100pt}
    \caption{The figure presents a comparison of different style edits based on various source image editing approaches, using different random seeds. Each method is evaluated with three different seeds. For each part, the top row displays the edited source image, while the two rows below show novel views generated from the edited model. GC=GaussCtrl, CN=ControlNet.}
    \label{fig:seed_comparisons}
\end{figure*}
\renewcommand{\arraystretch}{1.0} 
\setlength{\tabcolsep}{1pt}

\renewcommand{\arraystretch}{1.5} 
\setlength{\tabcolsep}{1pt}
\begin{figure*}[ht]
    \centering
    \begin{adjustbox}{max width=\textwidth}
    \begin{tabular}{p{0.17\textwidth}p{0.17\textwidth}p{0.17\textwidth}p{0.17\textwidth}}
    \multicolumn{1}{c}{\textbf{Source View}} & 
    \multicolumn{1}{c}{\textbf{View 1}} & 
    \multicolumn{1}{c}{\textbf{View 2}} & 
    \multicolumn{1}{c}{\textbf{View 3}} \\

    \begin{minipage}[t][0.15\textheight][t]{\linewidth}
        \centering
        \includegraphics[width=0.97\linewidth]{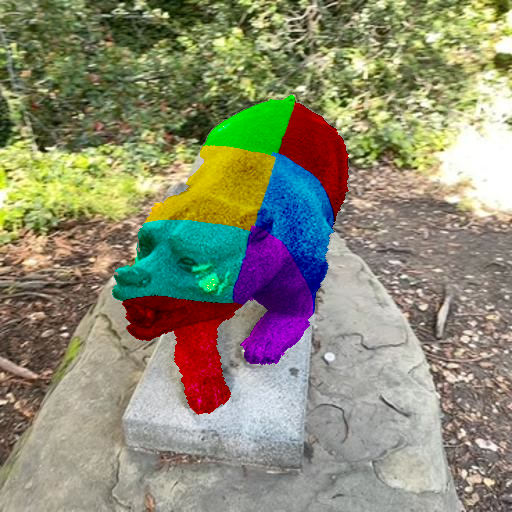}
    \end{minipage} &
    \begin{minipage}[t][0.15\textheight][t]{\linewidth}
        \centering
        \includegraphics[width=0.97\linewidth]{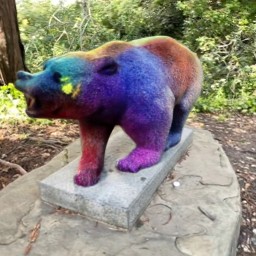}
    \end{minipage} &
    \begin{minipage}[t][0.15\textheight][t]{\linewidth}
        \centering
        \includegraphics[width=0.97\linewidth]{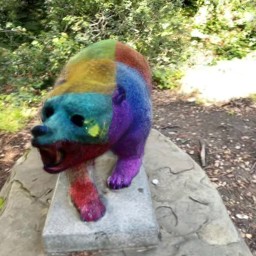}
    \end{minipage} &
    \begin{minipage}[t][0.15\textheight][t]{\linewidth}
        \centering
        \includegraphics[width=0.97\linewidth]{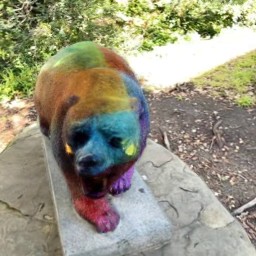}
    \end{minipage} \\

    \begin{minipage}[t][0.15\textheight][t]{\linewidth}
        \centering
        \includegraphics[width=0.97\linewidth]{supp_inversion_bear_bear_frame_00013_edit_3.png}
    \end{minipage} &
    \begin{minipage}[t][0.15\textheight][t]{\linewidth}
        \centering
        \includegraphics[width=0.97\linewidth]{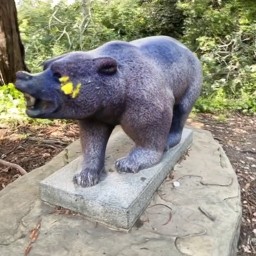}
    \end{minipage} &
    \begin{minipage}[t][0.15\textheight][t]{\linewidth}
        \centering
        \includegraphics[width=0.97\linewidth]{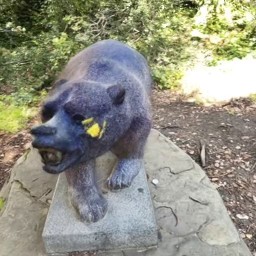}
    \end{minipage} &
    \begin{minipage}[t][0.15\textheight][t]{\linewidth}
        \centering
        \includegraphics[width=0.97\linewidth]{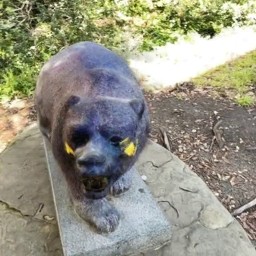}
    \end{minipage} \\

    \end{tabular}
    \end{adjustbox}
    \vspace{10pt}
    \caption{User-generated edits comparison between our method and the DGE method. The first row shows the user-provided edited image followed by three novel views generated using our method. The second row displays the same using the DGE method. This comparison highlights the differences in edit quality and consistency between the two approaches.}
    \label{fig:ddpm_inversion}

\end{figure*}
\renewcommand{\arraystretch}{1.0} 
\setlength{\tabcolsep}{1pt}

\begin{figure*}[!htbp]
    \centering
    \includegraphics[width=0.9\linewidth]{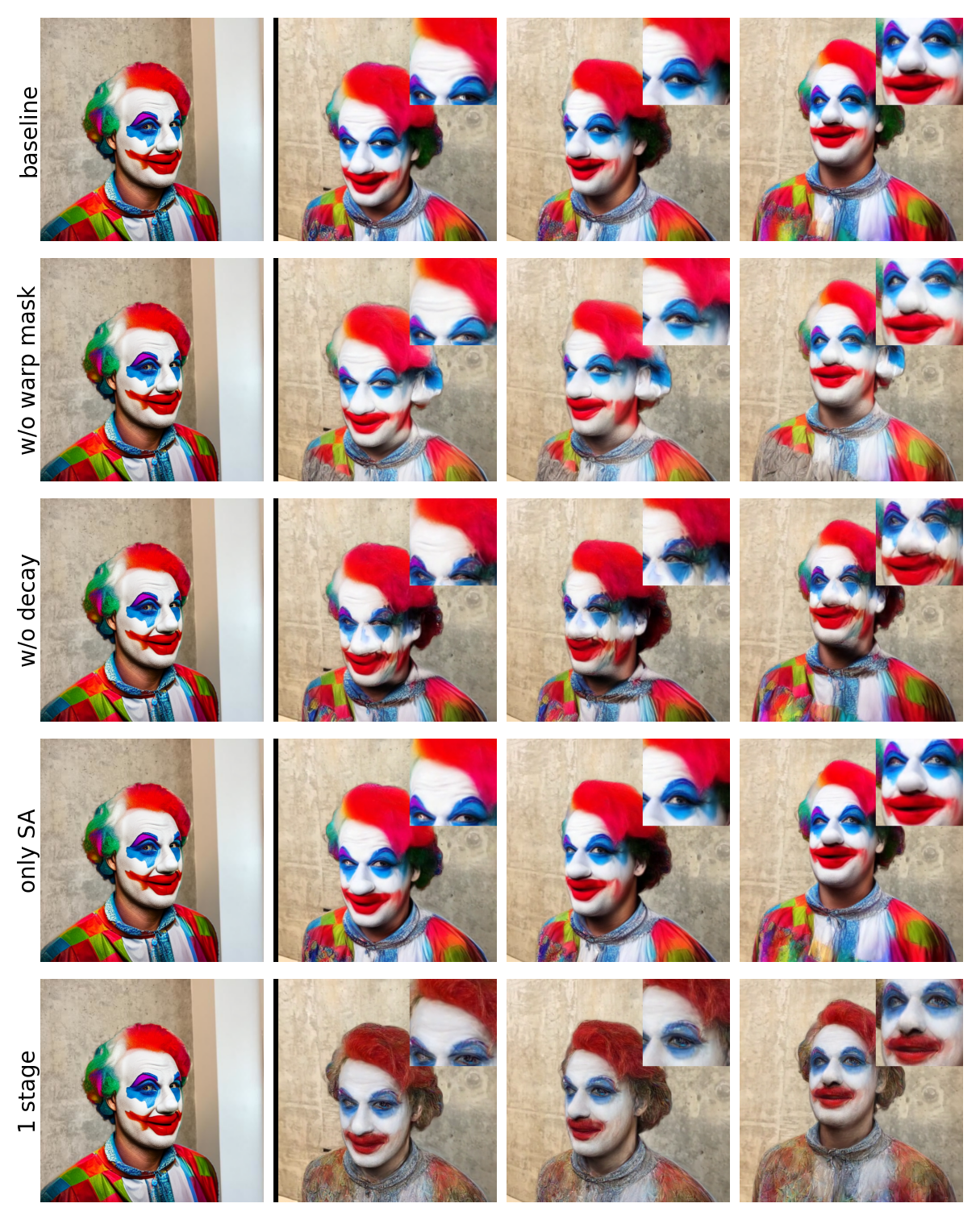}
    \caption{Ablation Study Overview. The figure shows how key components—warping mask, blending decay, attention types, and stage count—affect performance. The leftmost column is the edited source image for reference, with each row highlighting their impact on output quality. The first row represents the baseline, which is our full method, showcasing the effectiveness of all components combined.}
    \label{fig:ablation}
\end{figure*}

\renewcommand{\arraystretch}{1.5} 
\setlength{\tabcolsep}{1pt}
\begin{figure*}[ht]
    \centering
    \captionsetup{justification=centering}
    \begin{adjustbox}{max width=\textwidth}
    \begin{tabular}{p{0.15\textwidth}p{0.17\textwidth}p{0.17\textwidth}p{0.17\textwidth}p{0.17\textwidth}p{0.17\textwidth}p{0.17\textwidth}p{0.17\textwidth}}
    \multicolumn{1}{c}{\textbf{Source}} & \multicolumn{1}{c}{\textbf{IGS2GS}} & \multicolumn{1}{c}{\textbf{DGE}} & \multicolumn{1}{c}{\textbf{Ours (IP2P)}} & \multicolumn{1}{c}{\textbf{GC}} & \multicolumn{1}{c}{\textbf{Ours (CN)}} & \multicolumn{1}{c}{\textbf{GC Random}} & \multicolumn{1}{c}{\textbf{Ours CN Random}} \\

    \begin{minipage}[t][0.25\textheight][t]{\linewidth}
        \centering
        \includegraphics[width=0.98\linewidth]{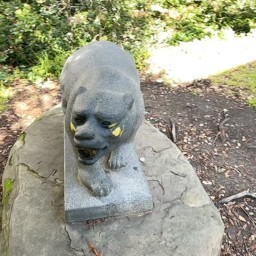}
    \end{minipage} &
    \begin{minipage}[t][0.22\textheight][t]{\linewidth}
        \centering
        \includegraphics[width=0.98\linewidth]{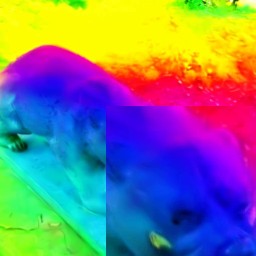} \\
        \vspace{3pt}
        \includegraphics[width=0.98\linewidth]{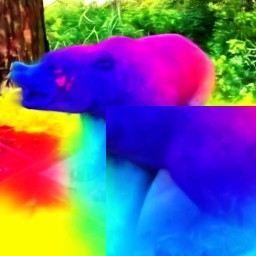}
    \end{minipage} &
    \begin{minipage}[t][0.22\textheight][t]{\linewidth}
        \centering
        \includegraphics[width=0.98\linewidth]{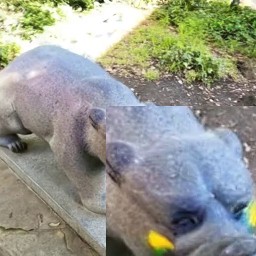} \\
        \vspace{3pt}
        \includegraphics[width=0.98\linewidth]{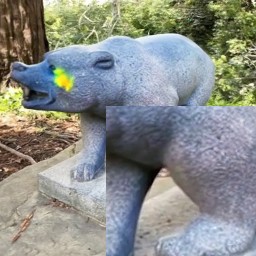}
    \end{minipage} &
    \begin{minipage}[t][0.22\textheight][t]{\linewidth}
        \centering
        \includegraphics[width=0.98\linewidth]{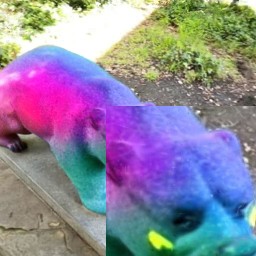} \\
        \vspace{3pt}
        \includegraphics[width=0.98\linewidth]{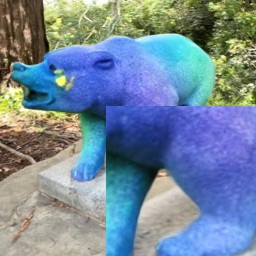}
    \end{minipage} &
    \begin{minipage}[t][0.22\textheight][t]{\linewidth}
        \centering
        \includegraphics[width=0.98\linewidth]{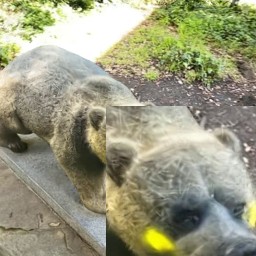} \\
        \vspace{3pt}
        \includegraphics[width=0.98\linewidth]{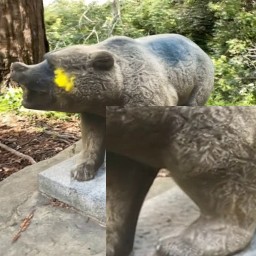}
    \end{minipage} &
    \begin{minipage}[t][0.22\textheight][t]{\linewidth}
        \centering
        \includegraphics[width=0.98\linewidth]{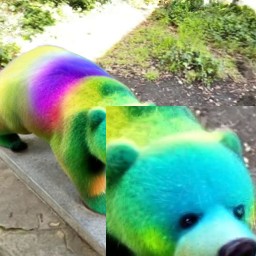} \\
        \vspace{3pt}
        \includegraphics[width=0.98\linewidth]{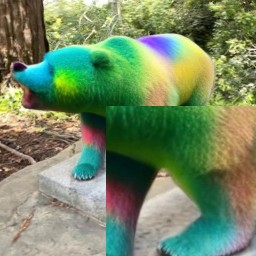}
    \end{minipage} &
    \begin{minipage}[t][0.22\textheight][t]{\linewidth}
        \centering
        \includegraphics[width=0.98\linewidth]{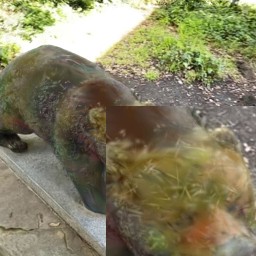} \\
        \vspace{3pt}
        \includegraphics[width=0.98\linewidth]{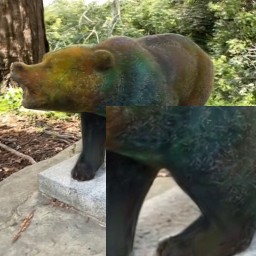}
    \end{minipage} &
    \begin{minipage}[t][0.22\textheight][t]{\linewidth}
        \centering
        \includegraphics[width=0.98\linewidth]{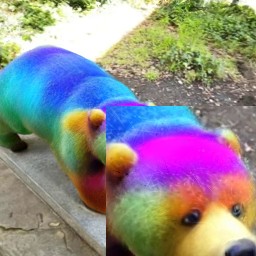} \\
        \vspace{3pt}
        \includegraphics[width=0.98\linewidth]{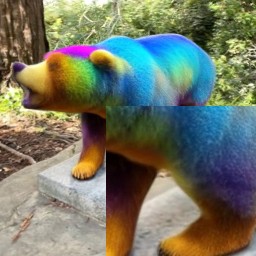}
    \end{minipage} \\

    \parbox[t]{\linewidth}{\centering \textbf{EDIT PSNR}} &
    \parbox[t]{\linewidth}{\centering 18.471} &
    \parbox[t]{\linewidth}{\centering \textbf{22.966}} &
    \parbox[t]{\linewidth}{\centering 22.840} &
    \parbox[t]{\linewidth}{\centering \textbf{27.020}} &
    \parbox[t]{\linewidth}{\centering 21.962} &
    \parbox[t]{\linewidth}{\centering \textbf{21.311}} &
    \parbox[t]{\linewidth}{\centering 20.224} \\

    \end{tabular}
    \end{adjustbox}
    \caption{This figure presents a comparison of Edit PSNR against edit quality. For each method, we show two different views along with a center zoom-in to enable a more precise evaluation of image quality. The left column present the source image.}

    \label{fig:psnr}
\end{figure*}

\renewcommand{\arraystretch}{1.5} 
\setlength{\tabcolsep}{1pt}
\begin{figure*}[ht]
    \centering
    \captionsetup{justification=centering}
    \begin{adjustbox}{max width=\textwidth}
    \begin{tabular}{p{0.15\textwidth}p{0.17\textwidth}p{0.17\textwidth}p{0.17\textwidth}p{0.17\textwidth}p{0.17\textwidth}p{0.17\textwidth}p{0.17\textwidth}}
    \multicolumn{1}{c}{\textbf{Source}} & \multicolumn{1}{c}{\textbf{IGS2GS}} & \multicolumn{1}{c}{\textbf{DGE}} & \multicolumn{1}{c}{\textbf{Ours (IP2P)}} & \multicolumn{1}{c}{\textbf{GC}} & \multicolumn{1}{c}{\textbf{Ours (CN)}} & \multicolumn{1}{c}{\textbf{GC Random}} & \multicolumn{1}{c}{\textbf{Ours CN Random}} \\

    \begin{minipage}[t][0.25\textheight][t]{\linewidth} 
        \centering
        \includegraphics[width=0.98\linewidth]{main_A_bear_with_rainbow_color_igs2gs_src.jpg}
        \rotatebox{90}{\parbox{0.12\textheight}{\textbf{A bear with rainbow color}}}
    \end{minipage} &
    \begin{minipage}[t][0.22\textheight][t]{\linewidth}
        \centering
        \includegraphics[width=0.98\linewidth]{main_A_bear_with_rainbow_color_igs2gs_v1.jpg} \\
        \vspace{3pt}
        \includegraphics[width=0.98\linewidth]{main_A_bear_with_rainbow_color_igs2gs_v2.jpg}
    \end{minipage} &
    \begin{minipage}[t][0.22\textheight][t]{\linewidth}
        \centering
        \includegraphics[width=0.98\linewidth]{main_A_bear_with_rainbow_color_dge_v1.jpg} \\
        \vspace{3pt}
        \includegraphics[width=0.98\linewidth]{main_A_bear_with_rainbow_color_dge_v2.jpg}
    \end{minipage} &
    \begin{minipage}[t][0.22\textheight][t]{\linewidth}
        \centering
        \includegraphics[width=0.98\linewidth]{main_A_bear_with_rainbow_color_ours_ip2p_v1.jpg} \\
        \vspace{3pt}
        \includegraphics[width=0.98\linewidth]{main_A_bear_with_rainbow_color_ours_ip2p_v2.jpg}
    \end{minipage} &
    \begin{minipage}[t][0.22\textheight][t]{\linewidth}
        \centering
        \includegraphics[width=0.98\linewidth]{main_A_bear_with_rainbow_color_gc_v1.jpg} \\
        \vspace{3pt}
        \includegraphics[width=0.98\linewidth]{main_A_bear_with_rainbow_color_gc_v2.jpg}
    \end{minipage} &
    \begin{minipage}[t][0.22\textheight][t]{\linewidth}
        \centering
        \includegraphics[width=0.98\linewidth]{main_A_bear_with_rainbow_color_ours_controlnet_v1.jpg} \\
        \vspace{3pt}
        \includegraphics[width=0.98\linewidth]{main_A_bear_with_rainbow_color_ours_controlnet_v2.jpg}
    \end{minipage} &
    \begin{minipage}[t][0.22\textheight][t]{\linewidth}
        \centering
        \includegraphics[width=0.98\linewidth]{main_A_bear_with_rainbow_color_gc_nl_v1.jpg} \\
        \vspace{3pt}
        \includegraphics[width=0.98\linewidth]{main_A_bear_with_rainbow_color_gc_nl_v2.jpg}
    \end{minipage} &
    \begin{minipage}[t][0.22\textheight][t]{\linewidth}
        \centering
        \includegraphics[width=0.98\linewidth]{main_A_bear_with_rainbow_color_ours_controlnet_nl_v1.jpg} \\
        \vspace{3pt}
        \includegraphics[width=0.98\linewidth]{main_A_bear_with_rainbow_color_ours_controlnet_nl_v2.jpg}
    \end{minipage} \\

    \begin{minipage}[t][0.25\textheight][t]{\linewidth} 
        \centering
        \includegraphics[width=0.98\linewidth]{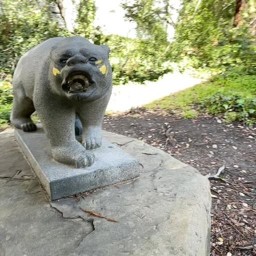}
        \rotatebox{90}{\parbox{0.12\textheight}{\textbf{A panda in the garden}}}
    \end{minipage} &
    \begin{minipage}[t][0.22\textheight][t]{\linewidth}
        \centering
        \includegraphics[width=0.98\linewidth]{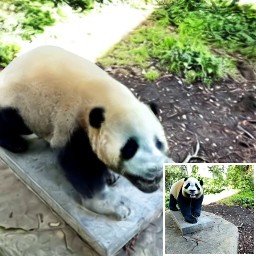} \\
        \vspace{3pt}
        \includegraphics[width=0.98\linewidth]{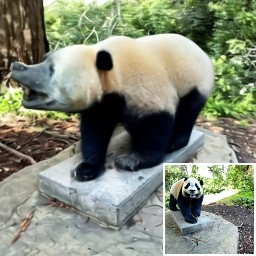}
    \end{minipage} &
    \begin{minipage}[t][0.22\textheight][t]{\linewidth}
        \centering
        \includegraphics[width=0.98\linewidth]{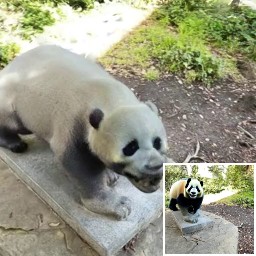} \\
        \vspace{3pt}
        \includegraphics[width=0.98\linewidth]{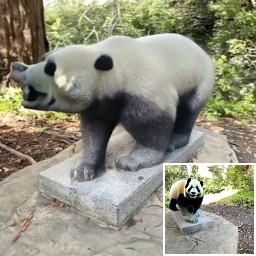}
    \end{minipage} &
    \begin{minipage}[t][0.22\textheight][t]{\linewidth}
        \centering
        \includegraphics[width=0.98\linewidth]{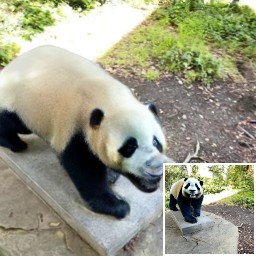} \\
        \vspace{3pt}
        \includegraphics[width=0.98\linewidth]{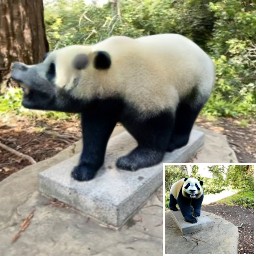}
    \end{minipage} &
    \begin{minipage}[t][0.22\textheight][t]{\linewidth}
        \centering
        \includegraphics[width=0.98\linewidth]{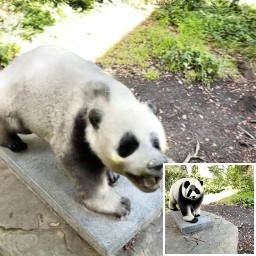} \\
        \vspace{3pt}
        \includegraphics[width=0.98\linewidth]{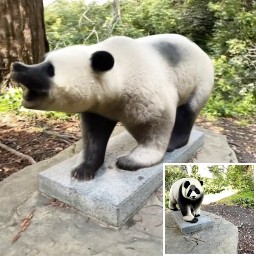}
    \end{minipage} &
    \begin{minipage}[t][0.22\textheight][t]{\linewidth}
        \centering
        \includegraphics[width=0.98\linewidth]{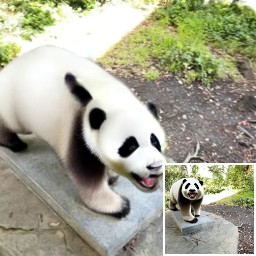} \\
        \vspace{3pt}
        \includegraphics[width=0.98\linewidth]{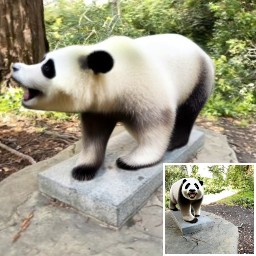}
    \end{minipage} &
    \begin{minipage}[t][0.22\textheight][t]{\linewidth}
        \centering
        \includegraphics[width=0.98\linewidth]{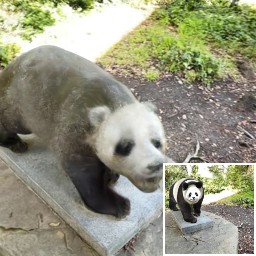} \\
        \vspace{3pt}
        \includegraphics[width=0.98\linewidth]{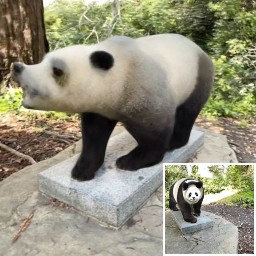}
    \end{minipage} &
    \begin{minipage}[t][0.22\textheight][t]{\linewidth}
        \centering
        \includegraphics[width=0.98\linewidth]{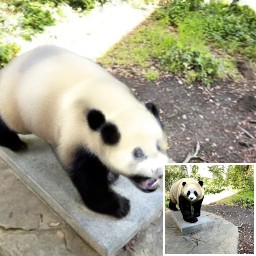} \\
        \vspace{3pt}
        \includegraphics[width=0.98\linewidth]{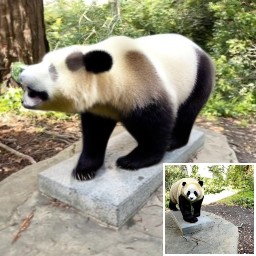}
    \end{minipage} \\

    \begin{minipage}[t][0.25\textheight][t]{\linewidth} 
        \centering
        \includegraphics[width=0.98\linewidth]{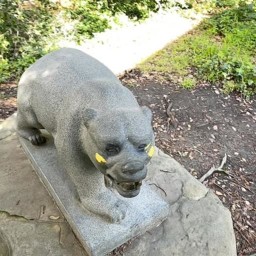}
        \rotatebox{90}{\parbox{0.12\textheight}{\textbf{A robotic bear in the garden}}}
    \end{minipage} &
    \begin{minipage}[t][0.22\textheight][t]{\linewidth}
        \centering
        \includegraphics[width=0.98\linewidth]{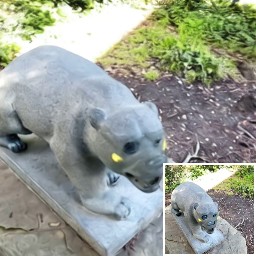} \\
        \vspace{3pt}
        \includegraphics[width=0.98\linewidth]{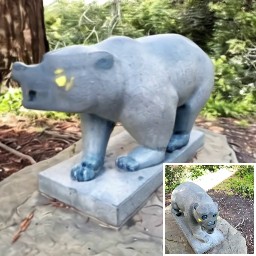}
    \end{minipage} &
    \begin{minipage}[t][0.22\textheight][t]{\linewidth}
        \centering
        \includegraphics[width=0.98\linewidth]{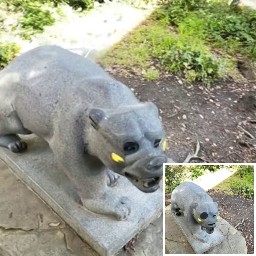} \\
        \vspace{3pt}
        \includegraphics[width=0.98\linewidth]{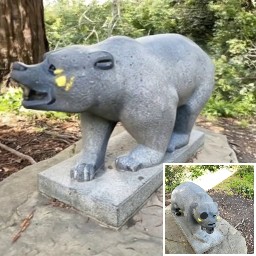}
    \end{minipage} &
    \begin{minipage}[t][0.22\textheight][t]{\linewidth}
        \centering
        \includegraphics[width=0.98\linewidth]{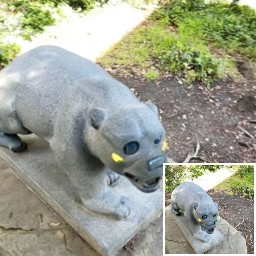} \\
        \vspace{3pt}
        \includegraphics[width=0.98\linewidth]{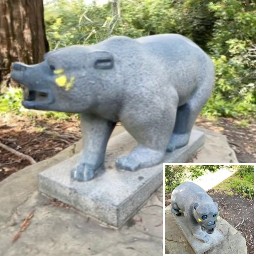}
    \end{minipage} &
    \begin{minipage}[t][0.22\textheight][t]{\linewidth}
        \centering
        \includegraphics[width=0.98\linewidth]{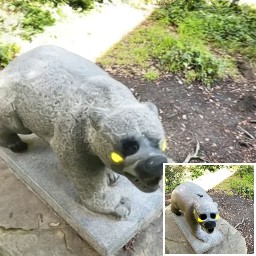} \\
        \vspace{3pt}
        \includegraphics[width=0.98\linewidth]{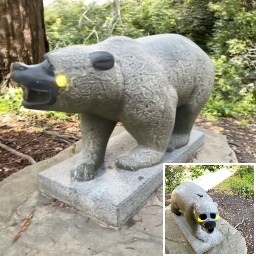}
    \end{minipage} &
    \begin{minipage}[t][0.22\textheight][t]{\linewidth}
        \centering
        \includegraphics[width=0.98\linewidth]{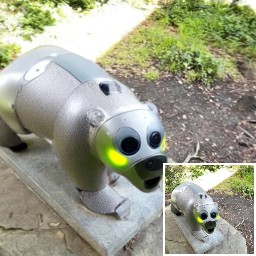} \\
        \vspace{3pt}
        \includegraphics[width=0.98\linewidth]{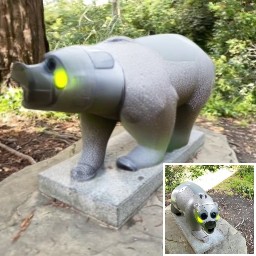}
    \end{minipage} &
    \begin{minipage}[t][0.22\textheight][t]{\linewidth}
        \centering
        \includegraphics[width=0.98\linewidth]{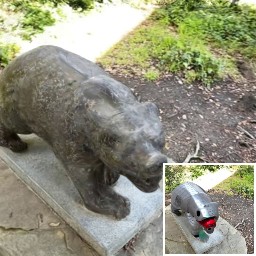} \\
        \vspace{3pt}
        \includegraphics[width=0.98\linewidth]{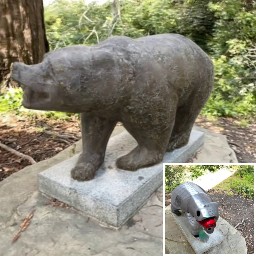}
    \end{minipage} &
    \begin{minipage}[t][0.22\textheight][t]{\linewidth}
        \centering
        \includegraphics[width=0.98\linewidth]{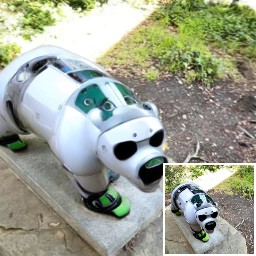} \\
        \vspace{3pt}
        \includegraphics[width=0.98\linewidth]{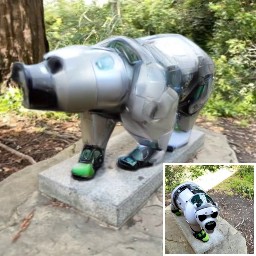}
    \end{minipage} \\

    \end{tabular}
    \end{adjustbox}
    \caption{A comparison of scene editing methods of \textit{bear} scene. Each sample shows two views, with the modified source image shown as an inset.}
    \label{fig:main_vis_bear}
\end{figure*}
\clearpage

\renewcommand{\arraystretch}{1.5} 
\setlength{\tabcolsep}{1pt}
\begin{figure*}[ht]
    \centering
    \captionsetup{justification=centering}
    \begin{adjustbox}{max width=\textwidth}
    \begin{tabular}{p{0.15\textwidth}p{0.17\textwidth}p{0.17\textwidth}p{0.17\textwidth}p{0.17\textwidth}p{0.17\textwidth}p{0.17\textwidth}p{0.17\textwidth}}
    \multicolumn{1}{c}{\textbf{Source}} & \multicolumn{1}{c}{\textbf{IGS2GS}} & \multicolumn{1}{c}{\textbf{DGE}} & \multicolumn{1}{c}{\textbf{Ours (IP2P)}} & \multicolumn{1}{c}{\textbf{GC}} & \multicolumn{1}{c}{\textbf{Ours (CN)}} & \multicolumn{1}{c}{\textbf{GC Random}} & \multicolumn{1}{c}{\textbf{Ours CN Random}} \\

    \begin{minipage}[t][0.25\textheight][t]{\linewidth} 
        \centering
        \includegraphics[width=0.98\linewidth]{main_A_dinosaur_statue_under_water_igs2gs_src.jpg}
        \rotatebox{90}{\parbox{0.12\textheight}{\textbf{A dinosaur statue under water}}}
    \end{minipage} &
    \begin{minipage}[t][0.22\textheight][t]{\linewidth}
        \centering
        \includegraphics[width=0.98\linewidth]{main_A_dinosaur_statue_under_water_igs2gs_v1.jpg} \\
        \vspace{3pt}
        \includegraphics[width=0.98\linewidth]{main_A_dinosaur_statue_under_water_igs2gs_v2.jpg}
    \end{minipage} &
    \begin{minipage}[t][0.22\textheight][t]{\linewidth}
        \centering
        \includegraphics[width=0.98\linewidth]{main_A_dinosaur_statue_under_water_dge_v1.jpg} \\
        \vspace{3pt}
        \includegraphics[width=0.98\linewidth]{main_A_dinosaur_statue_under_water_dge_v2.jpg}
    \end{minipage} &
    \begin{minipage}[t][0.22\textheight][t]{\linewidth}
        \centering
        \includegraphics[width=0.98\linewidth]{main_A_dinosaur_statue_under_water_ours_ip2p_v1.jpg} \\
        \vspace{3pt}
        \includegraphics[width=0.98\linewidth]{main_A_dinosaur_statue_under_water_ours_ip2p_v2.jpg}
    \end{minipage} &
    \begin{minipage}[t][0.22\textheight][t]{\linewidth}
        \centering
        \includegraphics[width=0.98\linewidth]{main_A_dinosaur_statue_under_water_gc_v1.jpg} \\
        \vspace{3pt}
        \includegraphics[width=0.98\linewidth]{main_A_dinosaur_statue_under_water_gc_v2.jpg}
    \end{minipage} &
    \begin{minipage}[t][0.22\textheight][t]{\linewidth}
        \centering
        \includegraphics[width=0.98\linewidth]{main_A_dinosaur_statue_under_water_ours_controlnet_v1.jpg} \\
        \vspace{3pt}
        \includegraphics[width=0.98\linewidth]{main_A_dinosaur_statue_under_water_ours_controlnet_v2.jpg}
    \end{minipage} &
    \begin{minipage}[t][0.22\textheight][t]{\linewidth}
        \centering
        \includegraphics[width=0.98\linewidth]{main_A_dinosaur_statue_under_water_gc_nl_v1.jpg} \\
        \vspace{3pt}
        \includegraphics[width=0.98\linewidth]{main_A_dinosaur_statue_under_water_gc_nl_v2.jpg}
    \end{minipage} &
    \begin{minipage}[t][0.22\textheight][t]{\linewidth}
        \centering
        \includegraphics[width=0.98\linewidth]{main_A_dinosaur_statue_under_water_ours_controlnet_nl_v1.jpg} \\
        \vspace{3pt}
        \includegraphics[width=0.98\linewidth]{main_A_dinosaur_statue_under_water_ours_controlnet_nl_v2.jpg}
    \end{minipage} \\

    \begin{minipage}[t][0.25\textheight][t]{\linewidth} 
        \centering
        \includegraphics[width=0.98\linewidth]{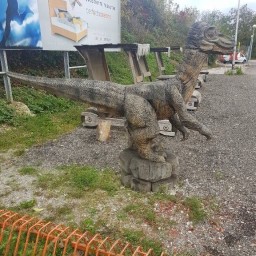}
        \rotatebox{90}{\parbox{0.12\textheight}{\textbf{A robot dinosaur on the road side}}}
    \end{minipage} &
    \begin{minipage}[t][0.22\textheight][t]{\linewidth}
        \centering
        \includegraphics[width=0.98\linewidth]{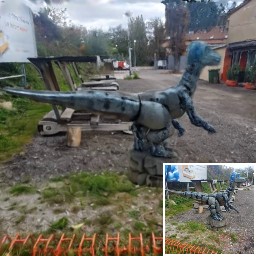} \\
        \vspace{3pt}
        \includegraphics[width=0.98\linewidth]{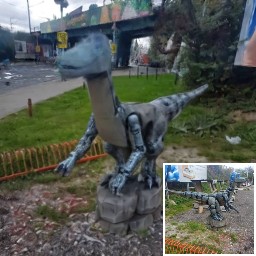}
    \end{minipage} &
    \begin{minipage}[t][0.22\textheight][t]{\linewidth}
        \centering
        \includegraphics[width=0.98\linewidth]{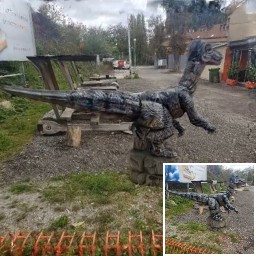} \\
        \vspace{3pt}
        \includegraphics[width=0.98\linewidth]{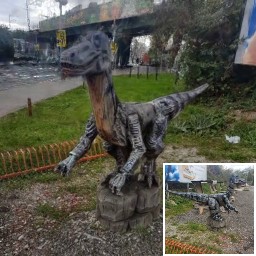}
    \end{minipage} &
    \begin{minipage}[t][0.22\textheight][t]{\linewidth}
        \centering
        \includegraphics[width=0.98\linewidth]{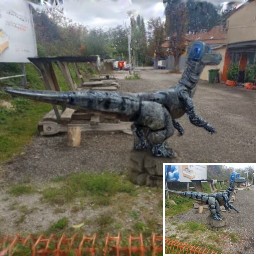} \\
        \vspace{3pt}
        \includegraphics[width=0.98\linewidth]{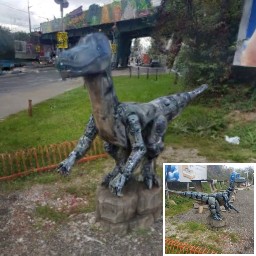}
    \end{minipage} &
    \begin{minipage}[t][0.22\textheight][t]{\linewidth}
        \centering
        \includegraphics[width=0.98\linewidth]{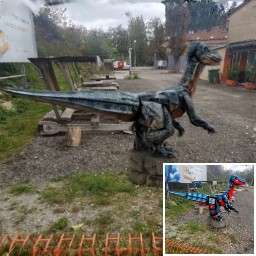} \\
        \vspace{3pt}
        \includegraphics[width=0.98\linewidth]{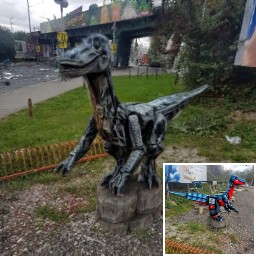}
    \end{minipage} &
    \begin{minipage}[t][0.22\textheight][t]{\linewidth}
        \centering
        \includegraphics[width=0.98\linewidth]{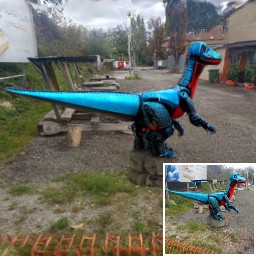} \\
        \vspace{3pt}
        \includegraphics[width=0.98\linewidth]{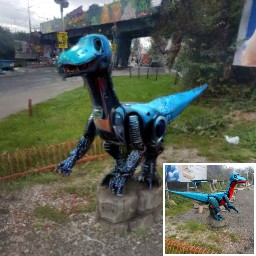}
    \end{minipage} &
    \begin{minipage}[t][0.22\textheight][t]{\linewidth}
        \centering
        \includegraphics[width=0.98\linewidth]{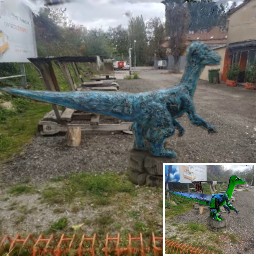} \\
        \vspace{3pt}
        \includegraphics[width=0.98\linewidth]{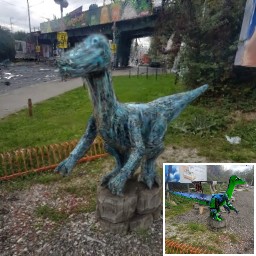}
    \end{minipage} &
    \begin{minipage}[t][0.22\textheight][t]{\linewidth}
        \centering
        \includegraphics[width=0.98\linewidth]{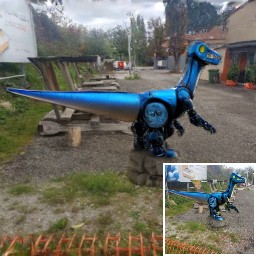} \\
        \vspace{3pt}
        \includegraphics[width=0.98\linewidth]{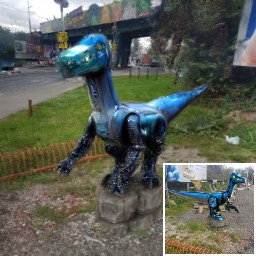}
    \end{minipage} \\

    \end{tabular}
    \end{adjustbox}
    \caption{A comparison of scene editing methods of \textit{dinosaur} scene. Each sample shows two views, with the modified source image shown as an inset.}
    \label{fig:main_vis_dino}
\end{figure*}
\clearpage

\renewcommand{\arraystretch}{1.5} 
\setlength{\tabcolsep}{1pt}
\begin{figure*}[ht]
    \centering
    \captionsetup{justification=centering}
    \begin{adjustbox}{max width=\textwidth}
    \begin{tabular}{p{0.15\textwidth}p{0.17\textwidth}p{0.17\textwidth}p{0.17\textwidth}p{0.17\textwidth}p{0.17\textwidth}p{0.17\textwidth}p{0.17\textwidth}}
    \multicolumn{1}{c}{\textbf{Source}} & \multicolumn{1}{c}{\textbf{IGS2GS}} & \multicolumn{1}{c}{\textbf{DGE}} & \multicolumn{1}{c}{\textbf{Ours (IP2P)}} & \multicolumn{1}{c}{\textbf{GC}} & \multicolumn{1}{c}{\textbf{Ours (CN)}} & \multicolumn{1}{c}{\textbf{GC Random}} & \multicolumn{1}{c}{\textbf{Ours CN Random}} \\

    \begin{minipage}[t][0.25\textheight][t]{\linewidth} 
        \centering
        \includegraphics[width=0.98\linewidth]{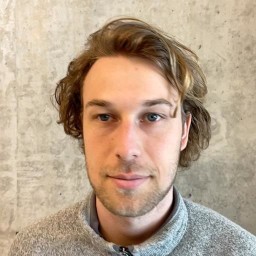}
        \rotatebox{90}{\parbox{0.12\textheight}{\textbf{A man with curly hair in a grey jacket with a Venetian mask}}}
    \end{minipage} &
    \begin{minipage}[t][0.22\textheight][t]{\linewidth}
        \centering
        \includegraphics[width=0.98\linewidth]{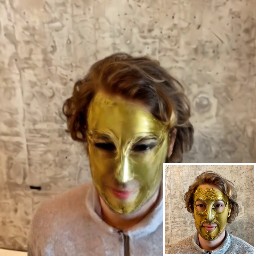} \\
        \vspace{3pt}
        \includegraphics[width=0.98\linewidth]{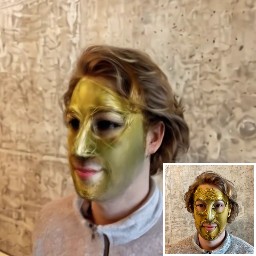}
    \end{minipage} &
    \begin{minipage}[t][0.22\textheight][t]{\linewidth}
        \centering
        \includegraphics[width=0.98\linewidth]{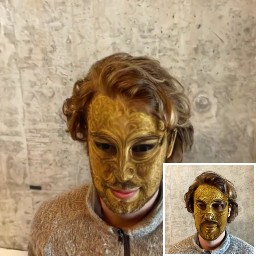} \\
        \vspace{3pt}
        \includegraphics[width=0.98\linewidth]{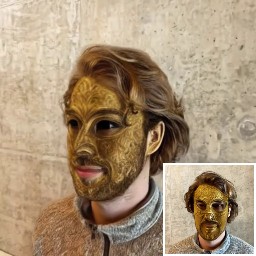}
    \end{minipage} &
    \begin{minipage}[t][0.22\textheight][t]{\linewidth}
        \centering
        \includegraphics[width=0.98\linewidth]{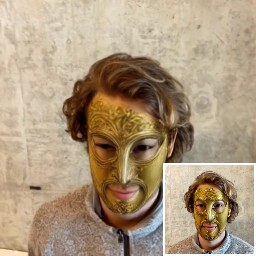} \\
        \vspace{3pt}
        \includegraphics[width=0.98\linewidth]{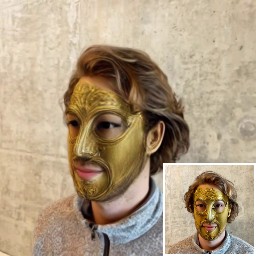}
    \end{minipage} &
    \begin{minipage}[t][0.22\textheight][t]{\linewidth}
        \centering
        \includegraphics[width=0.98\linewidth]{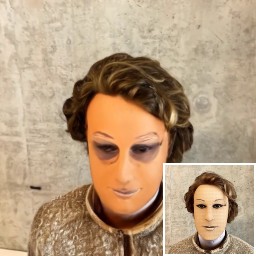} \\
        \vspace{3pt}
        \includegraphics[width=0.98\linewidth]{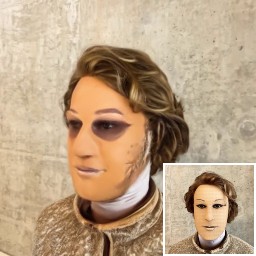}
    \end{minipage} &
    \begin{minipage}[t][0.22\textheight][t]{\linewidth}
        \centering
        \includegraphics[width=0.98\linewidth]{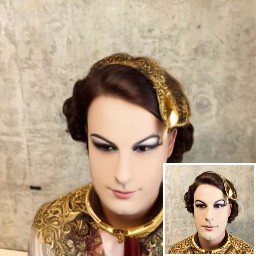} \\
        \vspace{3pt}
        \includegraphics[width=0.98\linewidth]{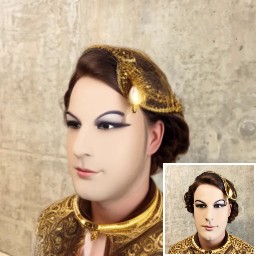}
    \end{minipage} &
    \begin{minipage}[t][0.22\textheight][t]{\linewidth}
        \centering
        \includegraphics[width=0.98\linewidth]{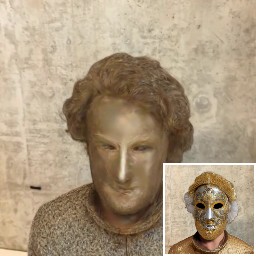} \\
        \vspace{3pt}
        \includegraphics[width=0.98\linewidth]{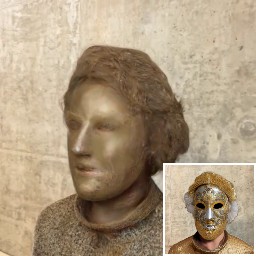}
    \end{minipage} &
    \begin{minipage}[t][0.22\textheight][t]{\linewidth}
        \centering
        \includegraphics[width=0.98\linewidth]{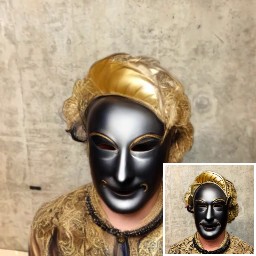} \\
        \vspace{3pt}
        \includegraphics[width=0.98\linewidth]{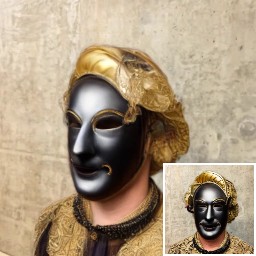}
    \end{minipage} \\

    \begin{minipage}[t][0.25\textheight][t]{\linewidth} 
        \centering
        \includegraphics[width=0.98\linewidth]{main_A_spider_man_with_a_mask_and_curly_hair_igs2gs_src.jpg}
        \rotatebox{90}{\parbox{0.12\textheight}{\textbf{A spider man with a mask and curly hair}}}
    \end{minipage} &
    \begin{minipage}[t][0.22\textheight][t]{\linewidth}
        \centering
        \includegraphics[width=0.98\linewidth]{main_A_spider_man_with_a_mask_and_curly_hair_igs2gs_v1.jpg} \\
        \vspace{3pt}
        \includegraphics[width=0.98\linewidth]{main_A_spider_man_with_a_mask_and_curly_hair_igs2gs_v2.jpg}
    \end{minipage} &
    \begin{minipage}[t][0.22\textheight][t]{\linewidth}
        \centering
        \includegraphics[width=0.98\linewidth]{main_A_spider_man_with_a_mask_and_curly_hair_dge_v1.jpg} \\
        \vspace{3pt}
        \includegraphics[width=0.98\linewidth]{main_A_spider_man_with_a_mask_and_curly_hair_dge_v2.jpg}
    \end{minipage} &
    \begin{minipage}[t][0.22\textheight][t]{\linewidth}
        \centering
        \includegraphics[width=0.98\linewidth]{main_A_spider_man_with_a_mask_and_curly_hair_ours_ip2p_v1.jpg} \\
        \vspace{3pt}
        \includegraphics[width=0.98\linewidth]{main_A_spider_man_with_a_mask_and_curly_hair_ours_ip2p_v2.jpg}
    \end{minipage} &
    \begin{minipage}[t][0.22\textheight][t]{\linewidth}
        \centering
        \includegraphics[width=0.98\linewidth]{main_A_spider_man_with_a_mask_and_curly_hair_gc_v1.jpg} \\
        \vspace{3pt}
        \includegraphics[width=0.98\linewidth]{main_A_spider_man_with_a_mask_and_curly_hair_gc_v2.jpg}
    \end{minipage} &
    \begin{minipage}[t][0.22\textheight][t]{\linewidth}
        \centering
        \includegraphics[width=0.98\linewidth]{main_A_spider_man_with_a_mask_and_curly_hair_ours_controlnet_v1.jpg} \\
        \vspace{3pt}
        \includegraphics[width=0.98\linewidth]{main_A_spider_man_with_a_mask_and_curly_hair_ours_controlnet_v2.jpg}
    \end{minipage} &
    \begin{minipage}[t][0.22\textheight][t]{\linewidth}
        \centering
        \includegraphics[width=0.98\linewidth]{main_A_spider_man_with_a_mask_and_curly_hair_gc_nl_v1.jpg} \\
        \vspace{3pt}
        \includegraphics[width=0.98\linewidth]{main_A_spider_man_with_a_mask_and_curly_hair_gc_nl_v2.jpg}
    \end{minipage} &
    \begin{minipage}[t][0.22\textheight][t]{\linewidth}
        \centering
        \includegraphics[width=0.98\linewidth]{main_A_spider_man_with_a_mask_and_curly_hair_ours_controlnet_nl_v1.jpg} \\
        \vspace{3pt}
        \includegraphics[width=0.98\linewidth]{main_A_spider_man_with_a_mask_and_curly_hair_ours_controlnet_nl_v2.jpg}
    \end{minipage} \\

    \begin{minipage}[t][0.25\textheight][t]{\linewidth} 
        \centering
        \includegraphics[width=0.98\linewidth]{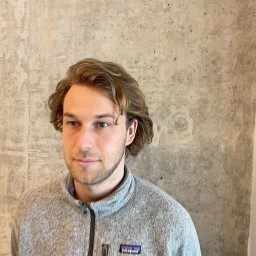}
        \rotatebox{90}{\parbox{0.12\textheight}{\textbf{A man with curly hair in a checkered cloth}}}
    \end{minipage} &
    \begin{minipage}[t][0.22\textheight][t]{\linewidth}
        \centering
        \includegraphics[width=0.98\linewidth]{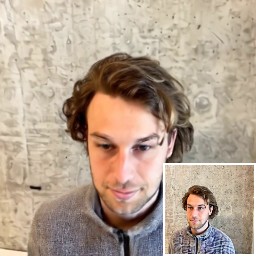} \\
        \vspace{3pt}
        \includegraphics[width=0.98\linewidth]{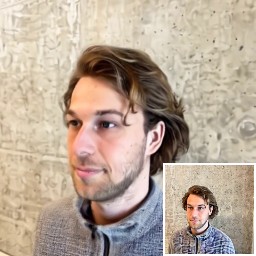}
    \end{minipage} &
    \begin{minipage}[t][0.22\textheight][t]{\linewidth}
        \centering
        \includegraphics[width=0.98\linewidth]{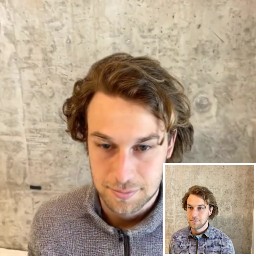} \\
        \vspace{3pt}
        \includegraphics[width=0.98\linewidth]{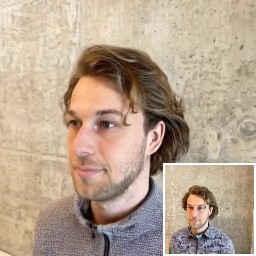}
    \end{minipage} &
    \begin{minipage}[t][0.22\textheight][t]{\linewidth}
        \centering
        \includegraphics[width=0.98\linewidth]{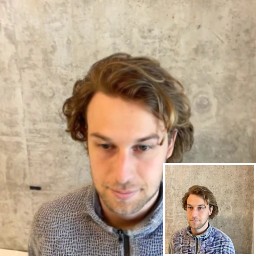} \\
        \vspace{3pt}
        \includegraphics[width=0.98\linewidth]{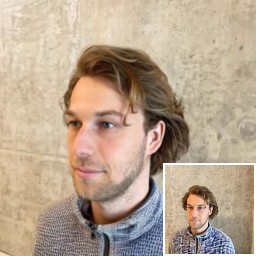}
    \end{minipage} &
    \begin{minipage}[t][0.22\textheight][t]{\linewidth}
        \centering
        \includegraphics[width=0.98\linewidth]{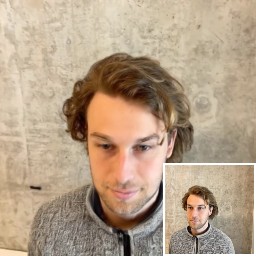} \\
        \vspace{3pt}
        \includegraphics[width=0.98\linewidth]{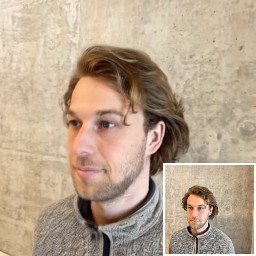}
    \end{minipage} &
    \begin{minipage}[t][0.22\textheight][t]{\linewidth}
        \centering
        \includegraphics[width=0.98\linewidth]{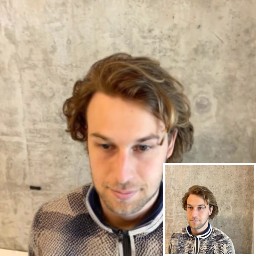} \\
        \vspace{3pt}
        \includegraphics[width=0.98\linewidth]{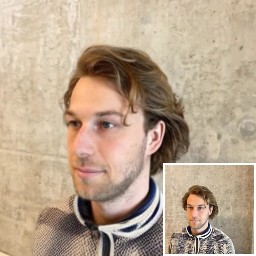}
    \end{minipage} &
    \begin{minipage}[t][0.22\textheight][t]{\linewidth}
        \centering
        \includegraphics[width=0.98\linewidth]{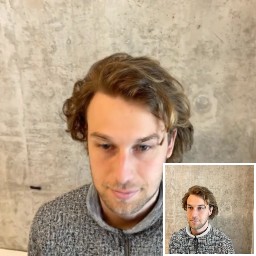} \\
        \vspace{3pt}
        \includegraphics[width=0.98\linewidth]{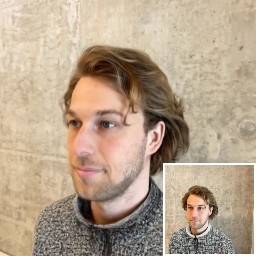}
    \end{minipage} &
    \begin{minipage}[t][0.22\textheight][t]{\linewidth}
        \centering
        \includegraphics[width=0.98\linewidth]{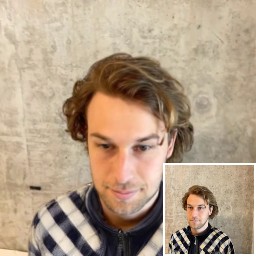} \\
        \vspace{3pt}
        \includegraphics[width=0.98\linewidth]{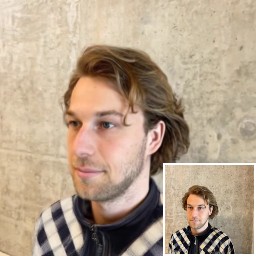}
    \end{minipage} \\

    \end{tabular}
    \end{adjustbox}
    \caption{A comparison of scene editing methods of \textit{face} scene. Each sample shows two views, with the modified source image shown as an inset.}
    \label{fig:main_vis_face}
\end{figure*}
\clearpage

\renewcommand{\arraystretch}{1.5} 
\setlength{\tabcolsep}{1pt}
\begin{figure*}[ht]
    \centering
    \captionsetup{justification=centering}
    \begin{adjustbox}{max width=\textwidth}
    \begin{tabular}{p{0.15\textwidth}p{0.17\textwidth}p{0.17\textwidth}p{0.17\textwidth}p{0.17\textwidth}p{0.17\textwidth}p{0.17\textwidth}p{0.17\textwidth}}
    \multicolumn{1}{c}{\textbf{Source}} & \multicolumn{1}{c}{\textbf{IGS2GS}} & \multicolumn{1}{c}{\textbf{DGE}} & \multicolumn{1}{c}{\textbf{Ours (IP2P)}} & \multicolumn{1}{c}{\textbf{GC}} & \multicolumn{1}{c}{\textbf{Ours (CN)}} & \multicolumn{1}{c}{\textbf{GC Random}} & \multicolumn{1}{c}{\textbf{Ours CN Random}} \\

    \begin{minipage}[t][0.25\textheight][t]{\linewidth} 
        \centering
        \includegraphics[width=0.98\linewidth]{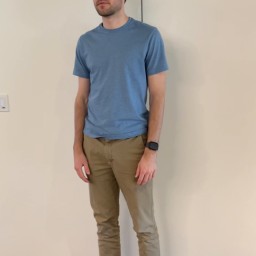}
        \rotatebox{90}{\parbox{0.12\textheight}{\textbf{A man wearing a shirt with a pineapple pattern}}}
    \end{minipage} &
    \begin{minipage}[t][0.22\textheight][t]{\linewidth}
        \centering
        \includegraphics[width=0.98\linewidth]{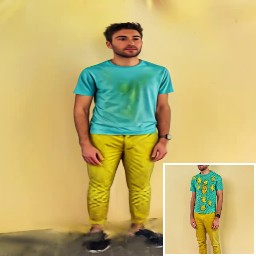} \\
        \vspace{3pt}
        \includegraphics[width=0.98\linewidth]{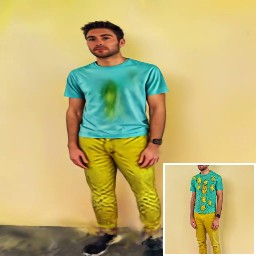}
    \end{minipage} &
    \begin{minipage}[t][0.22\textheight][t]{\linewidth}
        \centering
        \includegraphics[width=0.98\linewidth]{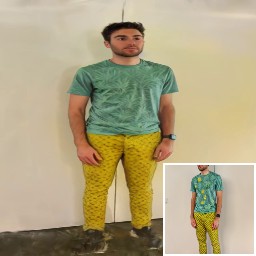} \\
        \vspace{3pt}
        \includegraphics[width=0.98\linewidth]{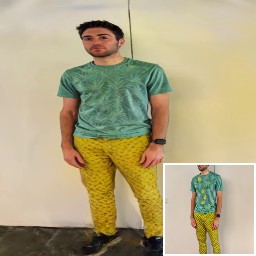}
    \end{minipage} &
    \begin{minipage}[t][0.22\textheight][t]{\linewidth}
        \centering
        \includegraphics[width=0.98\linewidth]{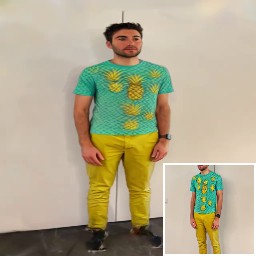} \\
        \vspace{3pt}
        \includegraphics[width=0.98\linewidth]{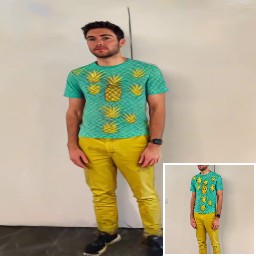}
    \end{minipage} &
    \begin{minipage}[t][0.22\textheight][t]{\linewidth}
        \centering
        \includegraphics[width=0.98\linewidth]{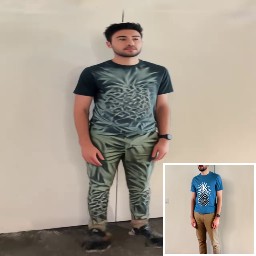} \\
        \vspace{3pt}
        \includegraphics[width=0.98\linewidth]{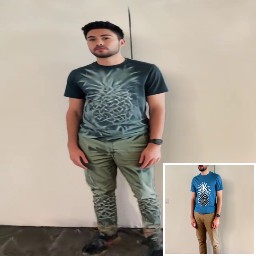}
    \end{minipage} &
    \begin{minipage}[t][0.22\textheight][t]{\linewidth}
        \centering
        \includegraphics[width=0.98\linewidth]{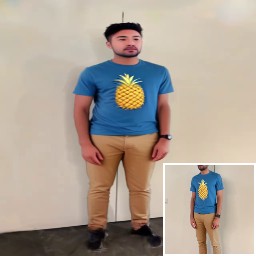} \\
        \vspace{3pt}
        \includegraphics[width=0.98\linewidth]{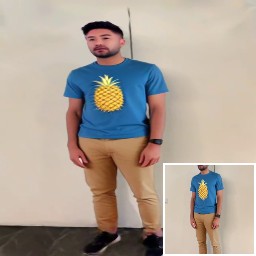}
    \end{minipage} &
    \begin{minipage}[t][0.22\textheight][t]{\linewidth}
        \centering
        \includegraphics[width=0.98\linewidth]{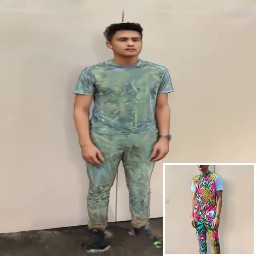} \\
        \vspace{3pt}
        \includegraphics[width=0.98\linewidth]{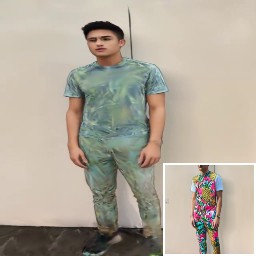}
    \end{minipage} &
    \begin{minipage}[t][0.22\textheight][t]{\linewidth}
        \centering
        \includegraphics[width=0.98\linewidth]{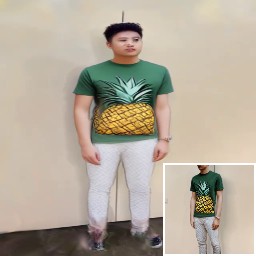} \\
        \vspace{3pt}
        \includegraphics[width=0.98\linewidth]{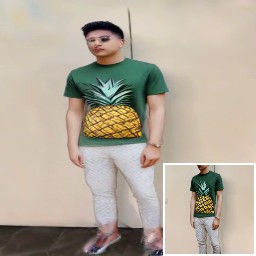}
    \end{minipage} \\

    \begin{minipage}[t][0.25\textheight][t]{\linewidth} 
        \centering
        \includegraphics[width=0.98\linewidth]{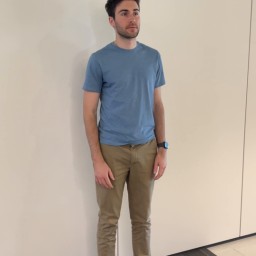}
        \rotatebox{90}{\parbox{0.12\textheight}{\textbf{An Iron Man stands to a wall}}}
    \end{minipage} &
    \begin{minipage}[t][0.22\textheight][t]{\linewidth}
        \centering
        \includegraphics[width=0.98\linewidth]{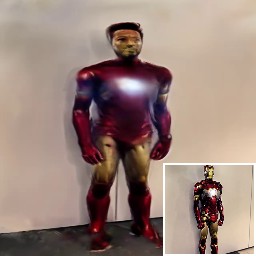} \\
        \vspace{3pt}
        \includegraphics[width=0.98\linewidth]{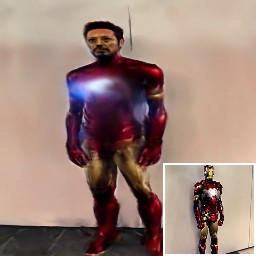}
    \end{minipage} &
    \begin{minipage}[t][0.22\textheight][t]{\linewidth}
        \centering
        \includegraphics[width=0.98\linewidth]{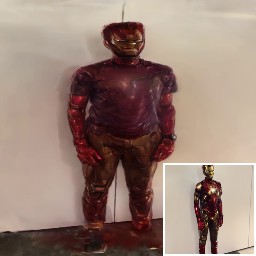} \\
        \vspace{3pt}
        \includegraphics[width=0.98\linewidth]{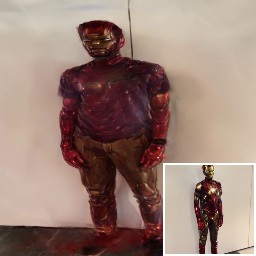}
    \end{minipage} &
    \begin{minipage}[t][0.22\textheight][t]{\linewidth}
        \centering
        \includegraphics[width=0.98\linewidth]{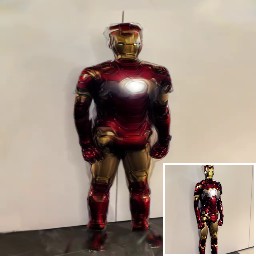} \\
        \vspace{3pt}
        \includegraphics[width=0.98\linewidth]{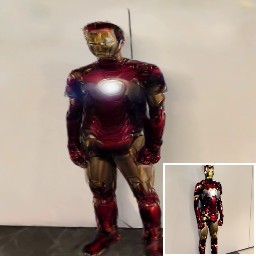}
    \end{minipage} &
    \begin{minipage}[t][0.22\textheight][t]{\linewidth}
        \centering
        \includegraphics[width=0.98\linewidth]{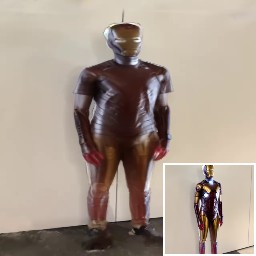} \\
        \vspace{3pt}
        \includegraphics[width=0.98\linewidth]{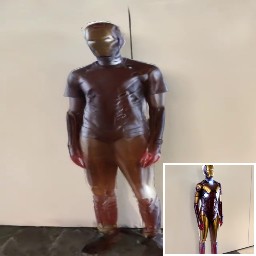}
    \end{minipage} &
    \begin{minipage}[t][0.22\textheight][t]{\linewidth}
        \centering
        \includegraphics[width=0.98\linewidth]{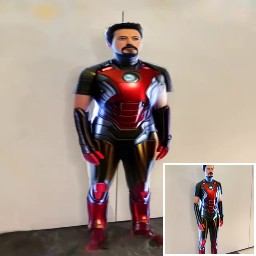} \\
        \vspace{3pt}
        \includegraphics[width=0.98\linewidth]{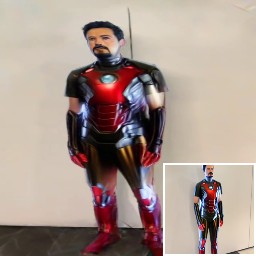}
    \end{minipage} &
    \begin{minipage}[t][0.22\textheight][t]{\linewidth}
        \centering
        \includegraphics[width=0.98\linewidth]{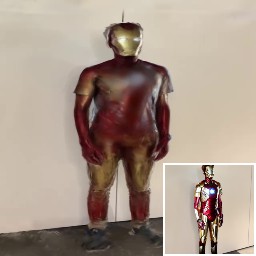} \\
        \vspace{3pt}
        \includegraphics[width=0.98\linewidth]{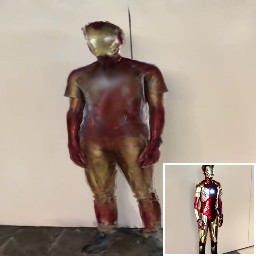}
    \end{minipage} &
    \begin{minipage}[t][0.22\textheight][t]{\linewidth}
        \centering
        \includegraphics[width=0.98\linewidth]{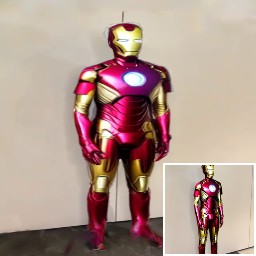} \\
        \vspace{3pt}
        \includegraphics[width=0.98\linewidth]{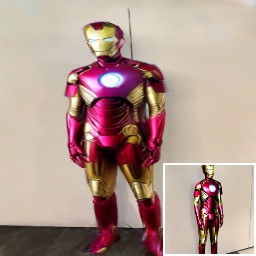}
    \end{minipage} \\

    \begin{minipage}[t][0.25\textheight][t]{\linewidth} 
        \centering
        \includegraphics[width=0.98\linewidth]{main_A_robot_stands_to_a_wall_igs2gs_src.jpg}
        \rotatebox{90}{\parbox{0.12\textheight}{\textbf{A robot stands to a wall}}}
    \end{minipage} &
    \begin{minipage}[t][0.22\textheight][t]{\linewidth}
        \centering
        \includegraphics[width=0.98\linewidth]{main_A_robot_stands_to_a_wall_igs2gs_v1.jpg} \\
        \vspace{3pt}
        \includegraphics[width=0.98\linewidth]{main_A_robot_stands_to_a_wall_igs2gs_v2.jpg}
    \end{minipage} &
    \begin{minipage}[t][0.22\textheight][t]{\linewidth}
        \centering
        \includegraphics[width=0.98\linewidth]{main_A_robot_stands_to_a_wall_dge_v1.jpg} \\
        \vspace{3pt}
        \includegraphics[width=0.98\linewidth]{main_A_robot_stands_to_a_wall_dge_v2.jpg}
    \end{minipage} &
    \begin{minipage}[t][0.22\textheight][t]{\linewidth}
        \centering
        \includegraphics[width=0.98\linewidth]{main_A_robot_stands_to_a_wall_ours_ip2p_v1.jpg} \\
        \vspace{3pt}
        \includegraphics[width=0.98\linewidth]{main_A_robot_stands_to_a_wall_ours_ip2p_v2.jpg}
    \end{minipage} &
    \begin{minipage}[t][0.22\textheight][t]{\linewidth}
        \centering
        \includegraphics[width=0.98\linewidth]{main_A_robot_stands_to_a_wall_gc_v1.jpg} \\
        \vspace{3pt}
        \includegraphics[width=0.98\linewidth]{main_A_robot_stands_to_a_wall_gc_v2.jpg}
    \end{minipage} &
    \begin{minipage}[t][0.22\textheight][t]{\linewidth}
        \centering
        \includegraphics[width=0.98\linewidth]{main_A_robot_stands_to_a_wall_ours_controlnet_v1.jpg} \\
        \vspace{3pt}
        \includegraphics[width=0.98\linewidth]{main_A_robot_stands_to_a_wall_ours_controlnet_v2.jpg}
    \end{minipage} &
    \begin{minipage}[t][0.22\textheight][t]{\linewidth}
        \centering
        \includegraphics[width=0.98\linewidth]{main_A_robot_stands_to_a_wall_gc_nl_v1.jpg} \\
        \vspace{3pt}
        \includegraphics[width=0.98\linewidth]{main_A_robot_stands_to_a_wall_gc_nl_v2.jpg}
    \end{minipage} &
    \begin{minipage}[t][0.22\textheight][t]{\linewidth}
        \centering
        \includegraphics[width=0.98\linewidth]{main_A_robot_stands_to_a_wall_ours_controlnet_nl_v1.jpg} \\
        \vspace{3pt}
        \includegraphics[width=0.98\linewidth]{main_A_robot_stands_to_a_wall_ours_controlnet_nl_v2.jpg}
    \end{minipage} \\

    \end{tabular}
    \end{adjustbox}
    \caption{A comparison of scene editing methods of \textit{person} scene. Each sample shows two views, with the modified source image shown as an inset.}
    \label{fig:main_vis_person}
\end{figure*}
\clearpage

\renewcommand{\arraystretch}{1.5} 
\setlength{\tabcolsep}{1pt}
\begin{figure*}[ht]
    \centering
    \captionsetup{justification=centering}
    \begin{adjustbox}{max width=\textwidth}
    \begin{tabular}{p{0.15\textwidth}p{0.17\textwidth}p{0.17\textwidth}p{0.17\textwidth}p{0.17\textwidth}p{0.17\textwidth}p{0.17\textwidth}p{0.17\textwidth}}
    \multicolumn{1}{c}{\textbf{Source}} & \multicolumn{1}{c}{\textbf{IGS2GS}} & \multicolumn{1}{c}{\textbf{DGE}} & \multicolumn{1}{c}{\textbf{Ours (IP2P)}} & \multicolumn{1}{c}{\textbf{GC}} & \multicolumn{1}{c}{\textbf{Ours (CN)}} & \multicolumn{1}{c}{\textbf{GC Random}} & \multicolumn{1}{c}{\textbf{Ours CN Random}} \\

    \begin{minipage}[t][0.25\textheight][t]{\linewidth} 
        \centering
        \includegraphics[width=0.98\linewidth]{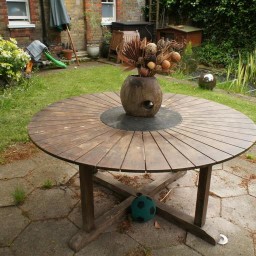}
        \rotatebox{90}{\parbox{0.12\textheight}{\textbf{A fake plant on a table in a garden covered with snow}}}
    \end{minipage} &
    \begin{minipage}[t][0.22\textheight][t]{\linewidth}
        \centering
        \includegraphics[width=0.98\linewidth]{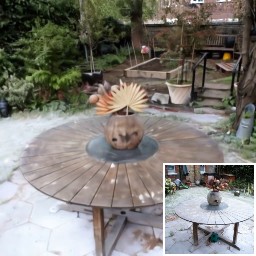} \\
        \vspace{3pt}
        \includegraphics[width=0.98\linewidth]{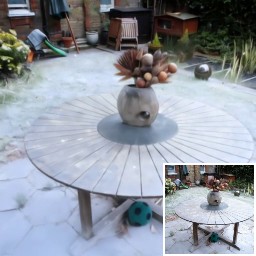}
    \end{minipage} &
    \begin{minipage}[t][0.22\textheight][t]{\linewidth}
        \centering
        \includegraphics[width=0.98\linewidth]{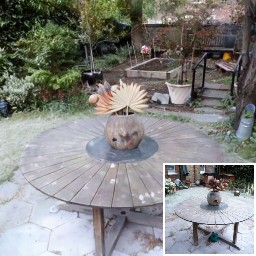} \\
        \vspace{3pt}
        \includegraphics[width=0.98\linewidth]{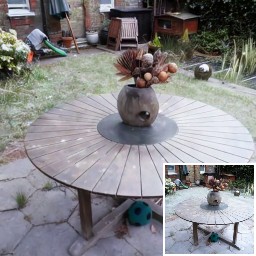}
    \end{minipage} &
    \begin{minipage}[t][0.22\textheight][t]{\linewidth}
        \centering
        \includegraphics[width=0.98\linewidth]{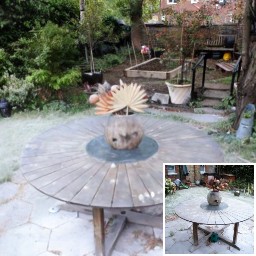} \\
        \vspace{3pt}
        \includegraphics[width=0.98\linewidth]{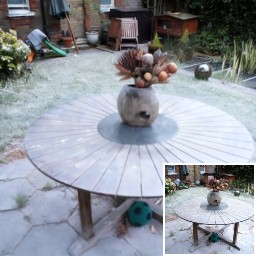}
    \end{minipage} &
    \begin{minipage}[t][0.22\textheight][t]{\linewidth}
        \centering
        \includegraphics[width=0.98\linewidth]{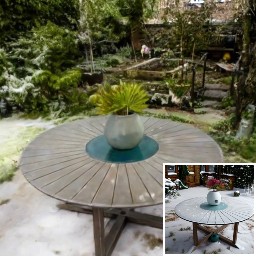} \\
        \vspace{3pt}
        \includegraphics[width=0.98\linewidth]{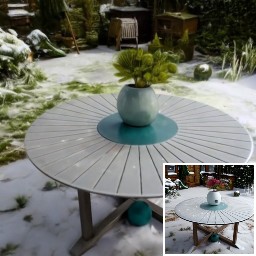}
    \end{minipage} &
    \begin{minipage}[t][0.22\textheight][t]{\linewidth}
        \centering
        \includegraphics[width=0.98\linewidth]{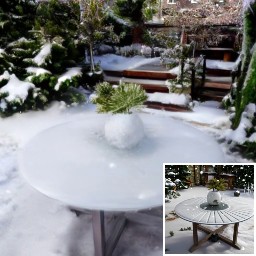} \\
        \vspace{3pt}
        \includegraphics[width=0.98\linewidth]{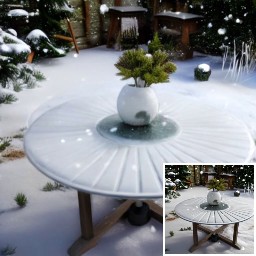}
    \end{minipage} &
    \begin{minipage}[t][0.22\textheight][t]{\linewidth}
        \centering
        \includegraphics[width=0.98\linewidth]{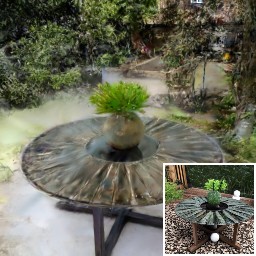} \\
        \vspace{3pt}
        \includegraphics[width=0.98\linewidth]{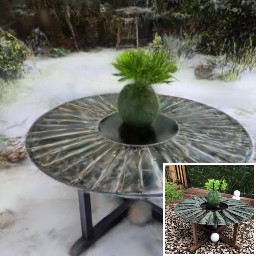}
    \end{minipage} &
    \begin{minipage}[t][0.22\textheight][t]{\linewidth}
        \centering
        \includegraphics[width=0.98\linewidth]{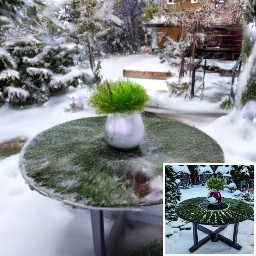} \\
        \vspace{3pt}
        \includegraphics[width=0.98\linewidth]{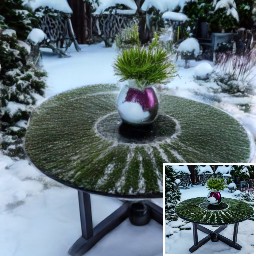}
    \end{minipage} \\

    \begin{minipage}[t][0.25\textheight][t]{\linewidth} 
        \centering
        \includegraphics[width=0.98\linewidth]{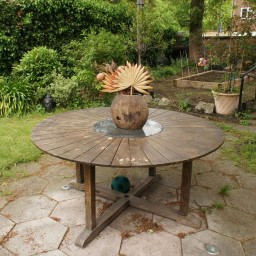}
        \rotatebox{90}{\parbox{0.12\textheight}{\textbf{A red fake plant on a table in the garden}}}
    \end{minipage} &
    \begin{minipage}[t][0.22\textheight][t]{\linewidth}
        \centering
        \includegraphics[width=0.98\linewidth]{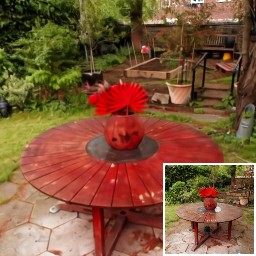} \\
        \vspace{3pt}
        \includegraphics[width=0.98\linewidth]{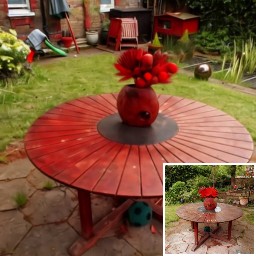}
    \end{minipage} &
    \begin{minipage}[t][0.22\textheight][t]{\linewidth}
        \centering
        \includegraphics[width=0.98\linewidth]{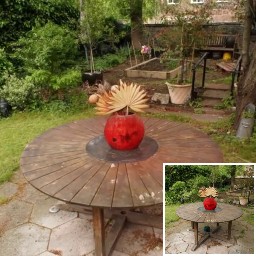} \\
        \vspace{3pt}
        \includegraphics[width=0.98\linewidth]{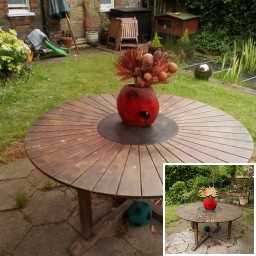}
    \end{minipage} &
    \begin{minipage}[t][0.22\textheight][t]{\linewidth}
        \centering
        \includegraphics[width=0.98\linewidth]{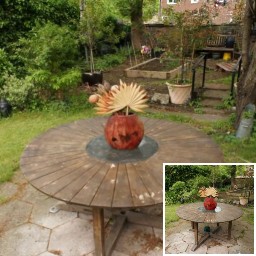} \\
        \vspace{3pt}
        \includegraphics[width=0.98\linewidth]{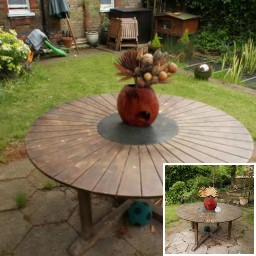}
    \end{minipage} &
    \begin{minipage}[t][0.22\textheight][t]{\linewidth}
        \centering
        \includegraphics[width=0.98\linewidth]{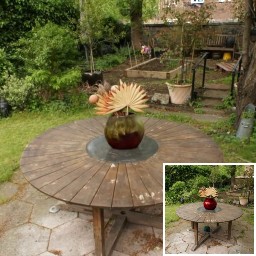} \\
        \vspace{3pt}
        \includegraphics[width=0.98\linewidth]{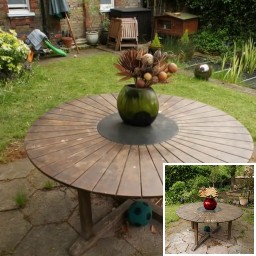}
    \end{minipage} &
    \begin{minipage}[t][0.22\textheight][t]{\linewidth}
        \centering
        \includegraphics[width=0.98\linewidth]{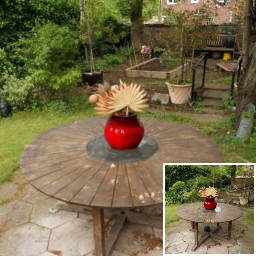} \\
        \vspace{3pt}
        \includegraphics[width=0.98\linewidth]{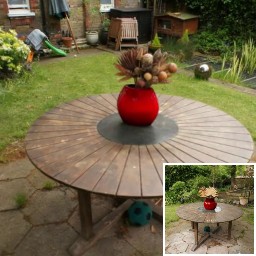}
    \end{minipage} &
    \begin{minipage}[t][0.22\textheight][t]{\linewidth}
        \centering
        \includegraphics[width=0.98\linewidth]{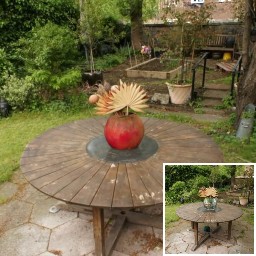} \\
        \vspace{3pt}
        \includegraphics[width=0.98\linewidth]{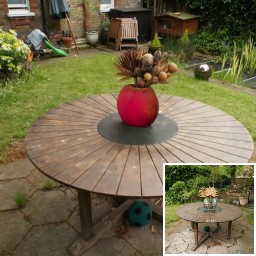}
    \end{minipage} &
    \begin{minipage}[t][0.22\textheight][t]{\linewidth}
        \centering
        \includegraphics[width=0.98\linewidth]{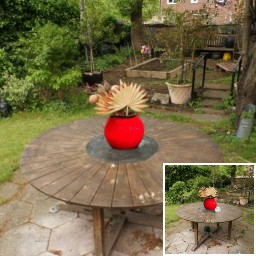} \\
        \vspace{3pt}
        \includegraphics[width=0.98\linewidth]{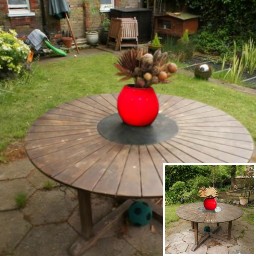}
    \end{minipage} \\

    \begin{minipage}[t][0.25\textheight][t]{\linewidth} 
        \centering
        \includegraphics[width=0.98\linewidth]{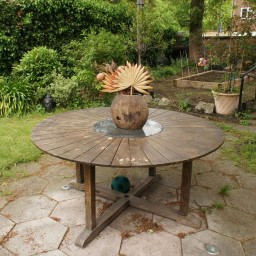}
        \rotatebox{90}{\parbox{0.12\textheight}{\textbf{A green fake plant on a table in the garden}}}
    \end{minipage} &
    \begin{minipage}[t][0.22\textheight][t]{\linewidth}
        \centering
        \includegraphics[width=0.98\linewidth]{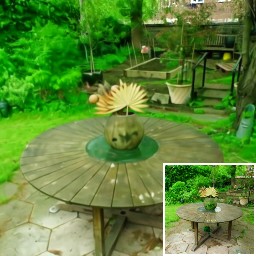} \\
        \vspace{3pt}
        \includegraphics[width=0.98\linewidth]{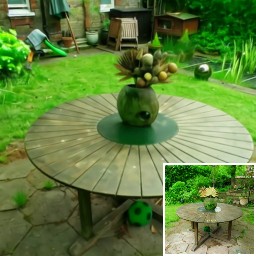}
    \end{minipage} &
    \begin{minipage}[t][0.22\textheight][t]{\linewidth}
        \centering
        \includegraphics[width=0.98\linewidth]{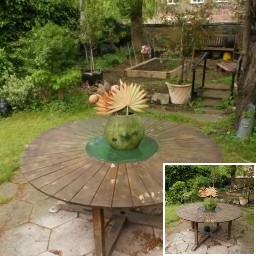} \\
        \vspace{3pt}
        \includegraphics[width=0.98\linewidth]{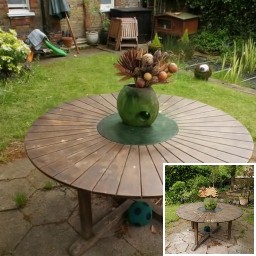}
    \end{minipage} &
    \begin{minipage}[t][0.22\textheight][t]{\linewidth}
        \centering
        \includegraphics[width=0.98\linewidth]{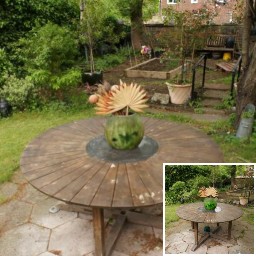} \\
        \vspace{3pt}
        \includegraphics[width=0.98\linewidth]{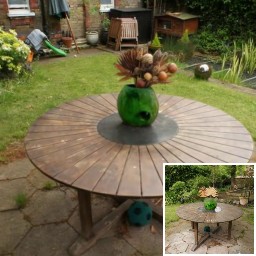}
    \end{minipage} &
    \begin{minipage}[t][0.22\textheight][t]{\linewidth}
        \centering
        \includegraphics[width=0.98\linewidth]{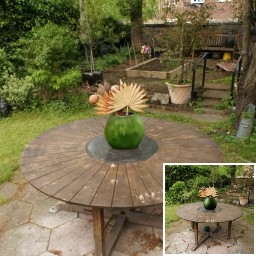} \\
        \vspace{3pt}
        \includegraphics[width=0.98\linewidth]{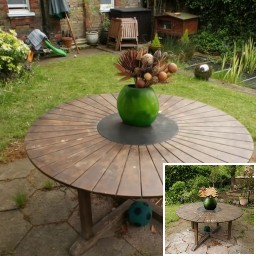}
    \end{minipage} &
    \begin{minipage}[t][0.22\textheight][t]{\linewidth}
        \centering
        \includegraphics[width=0.98\linewidth]{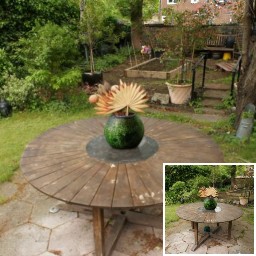} \\
        \vspace{3pt}
        \includegraphics[width=0.98\linewidth]{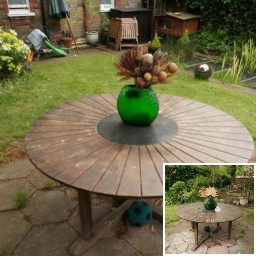}
    \end{minipage} &
    \begin{minipage}[t][0.22\textheight][t]{\linewidth}
        \centering
        \includegraphics[width=0.98\linewidth]{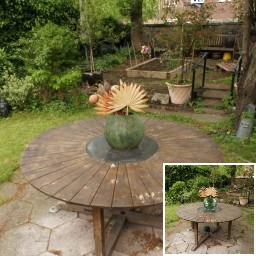} \\
        \vspace{3pt}
        \includegraphics[width=0.98\linewidth]{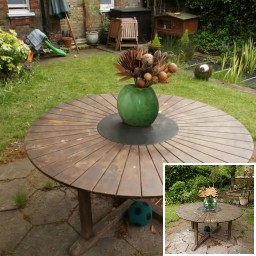}
    \end{minipage} &
    \begin{minipage}[t][0.22\textheight][t]{\linewidth}
        \centering
        \includegraphics[width=0.98\linewidth]{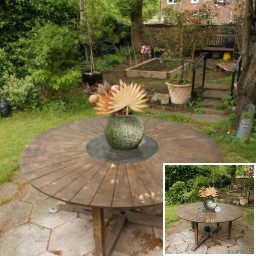} \\
        \vspace{3pt}
        \includegraphics[width=0.98\linewidth]{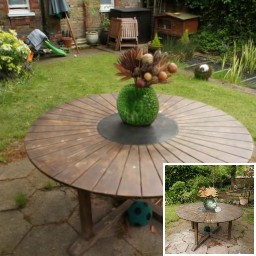}
    \end{minipage} \\

    \end{tabular}
    \end{adjustbox}
    \caption{A comparison of scene editing methods of \textit{table} scene. Each sample shows two views, with the modified source image shown as an inset.}
    \label{fig:main_vis_table}

\end{figure*}
\clearpage

\begin{table*}[ht]
\centering
\renewcommand{\arraystretch}{1.3}
\setlength{\tabcolsep}{5pt}
\begin{tabular}{|l|p{3.5cm}|p{3.5cm}|p{3.5cm}|p{3.5cm}|}
\hline
\textbf{Scene} & \textbf{Source Prompt} & \textbf{Target Prompt} & \textbf{Edit Prompt (IP2P)} & \textbf{Edit Prompt (CN)} \\ \hline
bear & A stone bear in a garden & A bear with rainbow color & Make the color of the bear look like rainbow & a photo a rainbow colored bear in the forest \\ \hline
bear & A stone bear in a garden & A robotic bear in the garden & Make the bear look like a robot & a photo of a robot bear in the forest \\ \hline
bear & A stone bear in a garden & A panda in the garden & Make the bear look like a panda & a photo of a panda bear in the forest \\ \hline
person & A man standing next to a wall wearing a blue T-shirt and brown pants & A man looks like a mosaic sculpture standing next to a wall & Make the man look like a mosaic sculpture & a photo of a man look like a mosaic sculpture \\ \hline
person & A man standing next to a wall wearing a blue T-shirt and brown pants & A robot stands to a wall & Turn the man into a robot & a photo of a robot \\ \hline
person & A man standing next to a wall wearing a blue T-shirt and brown pants & An Iron Man stands to a wall & Turn him into Iron Man & a photo of Iron Man \\ \hline
person & A man standing next to a wall wearing a blue T-shirt and brown pants & A man wearing a shirt with a pineapple pattern & Make the person wear a shirt with a pineapple pattern & a photo of a person wear a shirt with a pineapple pattern \\ \hline
face & A man with curly hair in a grey jacket & A man with curly hair in a grey jacket with a Venetian mask & Give him a Venetian mask & A man wearing a Venetian mask over his face \\ \hline
face & A man with curly hair in a grey jacket & A man with curly hair in a checkered cloth & Give him a checkered jacket & a photo of a man wearing a checkered jacket \\ \hline
face & A man with curly hair in a grey jacket & A spider man with a mask and curly hair & Turn him into spiderman with a mask & a photo of man wearing a Spiderman mask over his face \\ \hline
garden & A fake plant on a table in the garden & A fake plant on a table in a garden covered with snow & Make it snowy & a photo of a fake plant on a table in the garden in the snow \\ \hline
garden & A fake plant on a table in the garden & A red fake plant on a table in the garden & Turn the vase into red & a photo of a red fake plant on a table in the garden \\ \hline
garden & A fake plant on a table in the garden & A green fake plant on a table in the garden & Turn the vase into green & a photo of a green fake plant on a table in the garden \\ \hline
stump & A stump in the forest & A stump in the forest as Monet's painting & Turn it into Monet's painting & a photo of a stump as Monet's painting \\ \hline
stump & A stump in the forest & A stump in the forest on fire & Make it burn with fire & a photo of a stump on fire \\ \hline
dinosaur & A dinosaur statue on the road side & A robot dinosaur on the road side & Turn the dinosaur into a robot & a photo of a robot dinosaur on the road side \\ \hline
dinosaur & A dinosaur statue on the road side & A dinosaur statue under water & Make it underwater & a photo of a dinosaur statue under the water \\ \hline
\end{tabular}
\caption{Prompts for different scenes, showing variations across Source Prompts, Target Prompts, and Edit Prompts for both IP2P and ControlNet.}
\label{tab:scene_prompts}
\end{table*}
\clearpage

\begin{figure}[ht]
    \centering
    \includegraphics[width=\linewidth]{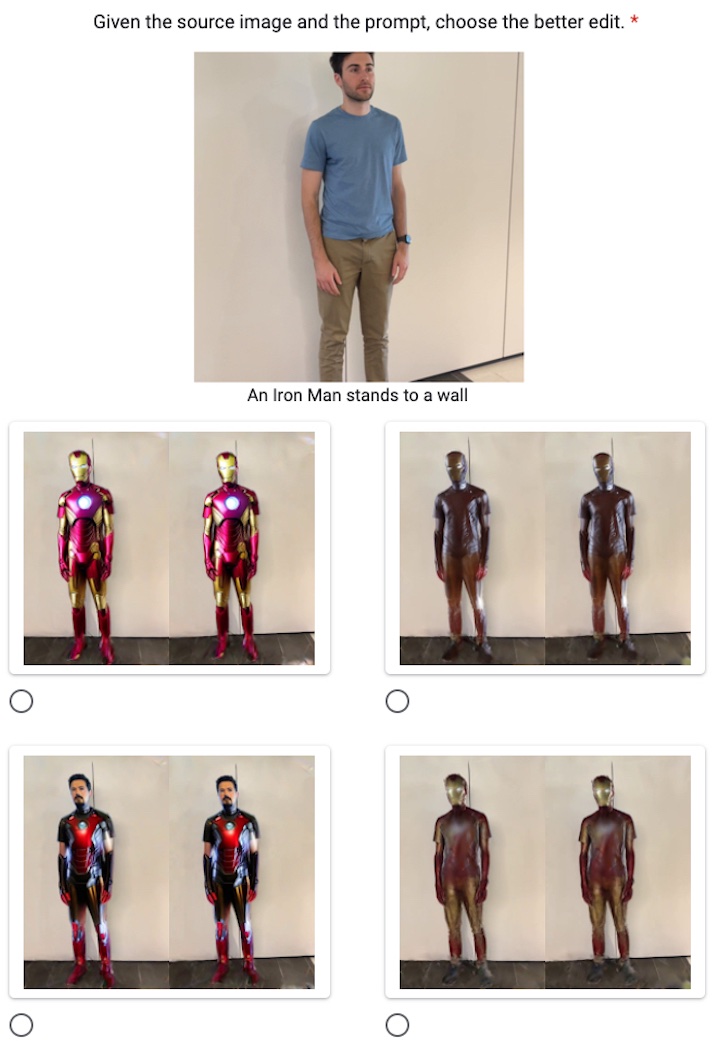}
    \caption{An example question from the user study designed to evaluate the subjective quality of edits. Participants were presented with results from multiple methods and asked to compare and select the edit that best adhered to the given prompt while maintaining visual fidelity.}
    \label{fig:user_study_example}
\end{figure}

\begin{figure}[ht]
    \centering
    \begin{minipage}{0.45\linewidth}
        \centering
        \includegraphics[width=\linewidth]{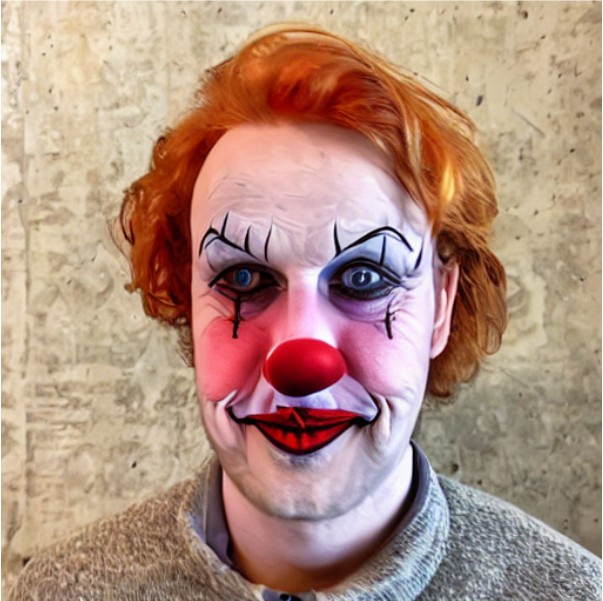}
        \subcaption{Source View}
    \end{minipage}
    \hfill
    \begin{minipage}{0.45\linewidth}
        \centering
        \includegraphics[width=\linewidth]{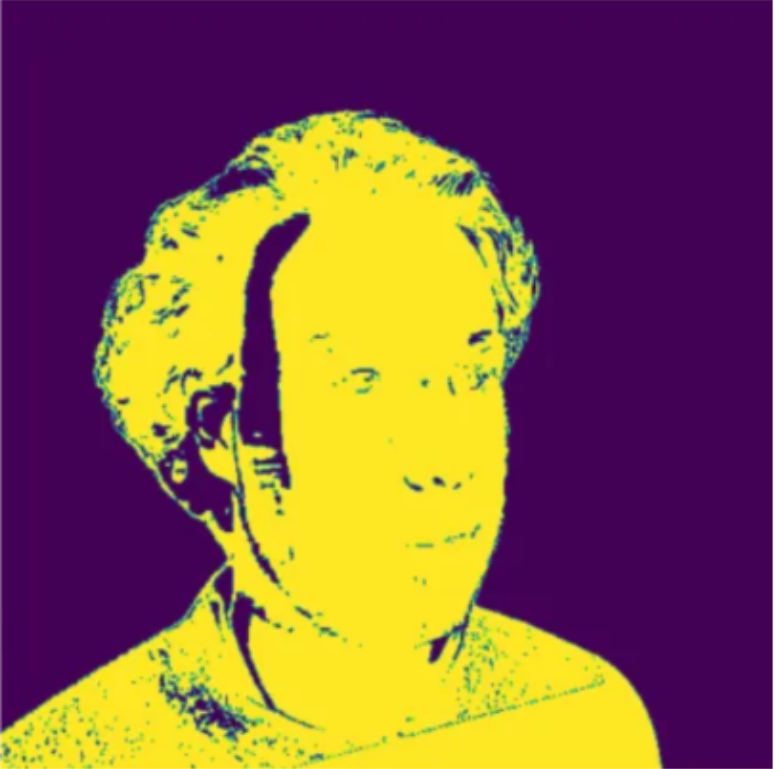}
        \subcaption{Validity Mask}
    \end{minipage}
    \vspace{5pt}
    \begin{minipage}{0.45\linewidth}
        \centering
        \includegraphics[width=\linewidth]{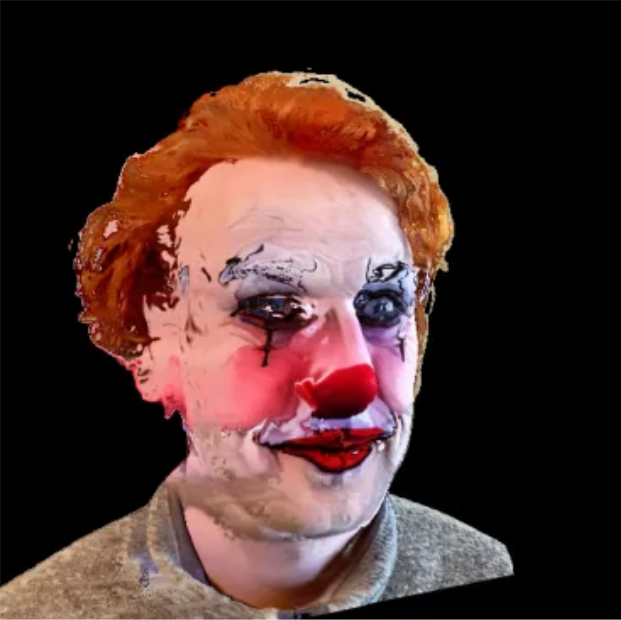}
        \subcaption{3DGS Warping}
    \end{minipage}
    \hfill
    \begin{minipage}{0.45\linewidth}
        \centering
        \includegraphics[width=\linewidth]{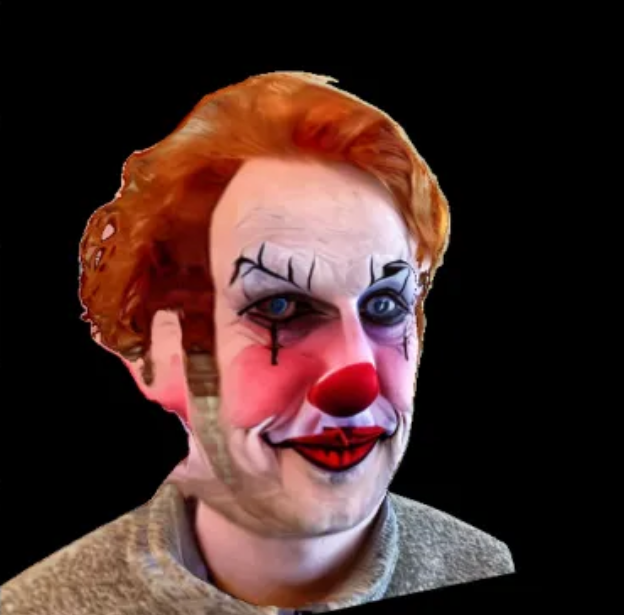}
        \subcaption{2DGS Warping}
    \end{minipage}
    \caption{A visual comparison of warping results using 2DGS~\cite{Huang2DGS2024} and 3DGS~\cite{kerbl3Dgaussians} depths. The top-left image shows the source view, the top-right image displays the warp validity mask, the bottom-left image presents warping using 3DGS, and the bottom-right image demonstrates warping using 2DGS. This figure highlights the superior geometric accuracy achieved with 2DGS, particularly in valid regions as indicated by the validity mask. Artifacts present in both methods can largely be attributed to areas outside the valid regions.}
    \label{fig:2d3dgs_warp}
\end{figure}

\end{document}